\documentclass{article}

     \usepackage[preprint]{neurips_data_2024}

\usepackage[utf8]{inputenc} 
\usepackage[T1]{fontenc}    
\usepackage{hyperref}       
\usepackage{url}            
\usepackage{booktabs}       
\usepackage{amsfonts}       
\usepackage{nicefrac}       
\usepackage{microtype}      
\usepackage{xcolor}         
\usepackage{graphicx}
\usepackage{wrapfig}

\usepackage{enumitem}
\setlist{topsep=0pt, leftmargin=*}

\usepackage[utf8]{inputenc} 
\usepackage[T1]{fontenc}    
\usepackage{hyperref}       
\usepackage{url}            
\usepackage{booktabs}       
\usepackage{amsfonts}       
\usepackage{nicefrac}       
\usepackage{microtype}      
\usepackage{xcolor}         
\usepackage{pdfpages}

\usepackage{cite}
\usepackage{url}    
\usepackage{xcolor}
\usepackage{wrapfig}
\usepackage{color, soul}
\usepackage{times}
\usepackage{colortbl}
\usepackage{amsmath}
\usepackage{makecell}
\usepackage{latexsym}
\usepackage{indentfirst}
\usepackage{enumerate}
\usepackage{multirow}
\usepackage{graphicx}
\usepackage{booktabs}
\usepackage{color}
\usepackage{listings}
\usepackage{amssymb}
\usepackage{inconsolata}
\usepackage{microtype}
\usepackage{textcomp}  
\usepackage{scalerel}  

\usepackage{caption,subcaption,graphicx}

\title{WikiContradict: A Benchmark for Evaluating LLMs on Real-World Knowledge Conflicts from Wikipedia}

\author{Yufang Hou$^{1}$, Alessandra Pascale$^{1}$, Javier Carnerero-Cano$^{1}$, Tigran Tchrakian$^{1}$ \\ \textbf{Radu Marinescu}$^{1}$, \textbf{Elizabeth Daly}$^{1}$, \textbf{Inkit Padhi}$^{2}$, \textbf{Prasanna Sattigeri}$^{2}$ \\
$^{1}$ IBM Research Europe - Ireland \\
$^{2}$ IBM Research, Thomas J. Watson Research Center, Yorktown Heights, USA\\
\texttt{\{yhou|apascale|tigran|radu.marinescu|elizabeth.daly\}@ie.ibm.com}   \\
\texttt{\{javier.cano|inkpad\}@ibm.com}, \texttt{psattig@us.ibm.com} 
}

\begin{document}

\maketitle

\begin{abstract}
Retrieval-augmented generation (RAG) has emerged as a promising solution to mitigate the limitations of large language models (LLMs), such as hallucinations and outdated information. However, it remains unclear how LLMs handle knowledge conflicts arising from different augmented retrieved passages, especially when these passages originate from the same source and have equal trustworthiness. In this work, we conduct a comprehensive evaluation of LLM-generated answers to questions that have varying answers based on contradictory passages from Wikipedia, a dataset widely regarded as a high-quality pre-training resource for most LLMs. Specifically, we introduce \texttt{WikiContradict}, a benchmark consisting of 253 high-quality, human-annotated instances designed to assess LLM performance when augmented with retrieved passages containing real-world knowledge conflicts. We benchmark a diverse range of both closed and open-source LLMs under different QA scenarios, including RAG with a single  passage, and RAG with 2 contradictory passages. Through rigorous human evaluations on a subset of \texttt{WikiContradict} instances involving 5 LLMs and over 3,500 judgements, we shed light on the behaviour and limitations of these models. For instance, when provided with two passages containing contradictory facts, all models struggle to generate answers that accurately reflect the conflicting nature of the context, especially for implicit conflicts requiring reasoning. Since human evaluation is costly, we
also introduce an automated model that estimates LLM performance using a strong open-source
language model, achieving an F-score of 0.8. Using this automated metric, we evaluate more than 1,500 answers from seven LLMs across all \texttt{WikiContradict} instances.
To facilitate future work, we release \texttt{WikiContradict} on: \url{https://ibm.biz/wikicontradict}.
\end{abstract}

\section{Introduction}
\label{sec:intro}

The advent of large language models (LLMs) \citep{LLMFewShortLearner} has revolutionized the field of Natural Language Processing (NLP), enabling unprecedented capabilities in text understanding and generation. However, static LLMs often suffer from outdated information and hallucinations. To mitigate these shortcomings, retrieval-augmented generation (RAG) techniques \citep{rag1} have been developed, which combine the strengths of LLMs with retrieved up-to-date information from external sources. 
While RAG frameworks have shown significant promise, it remains unclear how LLMs handle knowledge conflicts from different sources, including ``\emph{context-memory conflicts}'', which refers to the retrieved context knowledge being in conflict
with the parametric knowledge (memory) encapsulated within the LLM’s parameters,  and ``\emph{inter-context conflicts}'', which refers to the contradictions among the retrieved passages \citep{xu2024knowledge}. Most prior research on LLM knowledge conflicts has concentrated on ``\emph{context-memory conflicts}'' and relied on artificially generated datasets, which employ various methods to create conflicting information. These approaches span from simple entity substitution, where an entity in a passage is replaced with another entity of the same type \citep{longpre-etal-2021-entity}, to more sophisticated techniques, such as instructing language models like ChatGPT 
to fabricate supporting evidence for counterfactual answers to a given question \citep{xie2024knowledgeconflict,jin-etal-2024-tug-war}. However, these artificially generated datasets primarily focus on explicit, surface-level contradictions, neglecting the complexity and nuance of real-world knowledge conflicts.

\begin{figure}
    \centering 
    \includegraphics[width=.9\textwidth]{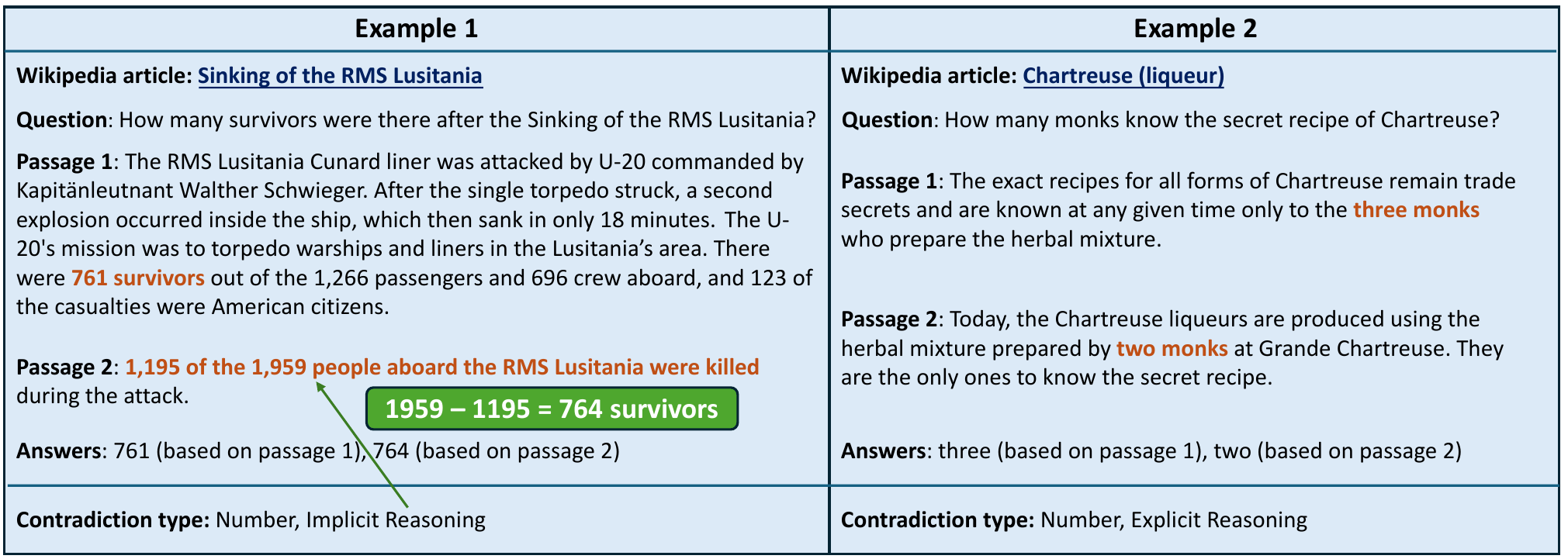}
    \caption{Example instances from \texttt{WikiContradict} with different contradiction types.}
    \label{fig:example}
\end{figure}

In this work, we focus on investigating the behaviors of LLMs when confronted with ``\textbf{\emph{real-world inter-context conflicts}}'', where knowledge inconsistencies arise from the same or different retrieved passages that originate from a single trusted source (Wikipedia) and are considered equally credible. Specifically, we introduce \texttt{WikiContradict}, a benchmark consisting of 253 high-quality, human-annotated instances that cover different types of contradictions identified by Wikipedia editors and validated by us.
Figure \ref{fig:example} presents two illustrative instances that demonstrate different types of contradictions. In particular, Example 1 requires implicit reasoning to detect the contradiction between Passage 1 and Passage 2 about the number of survivors from the RMS Lusitania sinking event, which requires calculating the number of survivors by subtracting 1,195 from 1,959 based on the information provided in Passage 2. This type of instance accounts for 36\% of the instances in the \texttt{WikiContradict} dataset.

We evaluate the performance of various LLMs on \texttt{WikiContradict} by employing diverse prompt templates to assess their behaviour under different question answering (QA) scenarios, including RAG with a single context passage, and RAG with two contradictory passages. We then conduct a rigorous human evaluation to assess the correctness of the models' responses. Our human evaluation dataset comprises responses from 5 LLMs to 5 prompt templates, applied to 55 instances from the \texttt{WikiContradict} dataset, resulting in a total of 1,375 evaluation samples. Each sample is annotated by 2 authors of this paper, yielding 2,750 human judgements. The inter-annotator agreement, measured by Cohen's $\kappa$, ranges from 0.58 to 0.88 across different prompt templates, indicating moderate to substantial agreement.
After resolving the annotation disagreements among annotators, our final human evaluation study dataset (\texttt{WikiContradict\_HumanEval}) consists of 1,200 samples resulting from 5 
LLMs' responses to 48 \texttt{WikiContradict} instances based on 5 prompt templates. 

On \texttt{WikiContradict\_HumanEval}, we observe that when instructing LLMs to generate answers to a given question based on the given context consisting of two contradicted passages, all
models, including GPT-4, struggle to generate answers that accurately reflect the conflicting nature of the context, especially for implicit conflicts that require reasoning as illustrated in Figure \ref {fig:example}, Example 1. Interestingly, we find that prompting LLMs to pay attention to contradictory context information improves their performance to correctly answering these questions. For instance, 
the top-performing model, Llama-3-70b-instruct, shows a remarkable increase from 10.4\% to 43.8\%. Furthermore, our analysis reveals that this improvement is largely driven by instances with explicit conflicts, as illustrated in Figure \ref{fig:example}, Example 2. Finally, to facilitate future evaluations, we have developed \texttt{WikiContradictEval}, a simple automatic evaluation method that leverages few-shot in-context learning to teach Llama-3-70b-instruct to judge model responses, which achieves an F-score of 0.8 on \texttt{WikiContradict\_HumanEval} for evaluating LLM responses in the RAG setting with two contradictory passages.

 In summary, our proposed \texttt{WikiContradict} benchmark poses a significant challenge for current LLMs, highlighting substantial opportunities for future improvement. We believe \texttt{WikiContradict} can serve as a valuable resource for the research community, facilitating the examination and tracking of LLMs' progress in handling real-world inter-context conflicts and deepening our understanding of their capabilities in these complex settings.

\section{Related work}
\label{sec:rel}
\textbf{Knowledge conflicts.~}
Knowledge conflicts are commonly presented to LLMs and exploring the capability of the model to understand and manage them to ensure trustworthiness of the answer is gaining increasing interest in the community. While there exist three categories of conflicts: intra-memory, context-memory and inter-context \citep{xu2024knowledge} we focus our attention on the inter-context conflict. This type of contradiction has become of particular interest after the advent of RAG techniques. RAG has been proven to enhance LLMs' capabilities in dealing with hallucination and enrich LLMs' responses by integrating content from retrieved documents into the context \citep{lewis2021retrievalaugmented}. At the same time RAG can also introduce inconsistencies, as external documents may conflict with each other.
In order to explore this phenomenon, previous research has relied on synthetically generated datasets containing conflicting statements \citep{wang2023resolving, li2024contradoc}. While they are used to evaluate and fine-tune existing models, these benchmarks fail to represent the complexity of real world conflicts \citep{xu2024knowledge}. We aim to shed light on the unexplored space of managing inter-context conflicts \textit{in the wild}, starting from contradictions extracted and annotated from Wikipedia. Our aim is to assess how well LLMs perform in dealing with real-world scenarios, rather than with synthetically created conflicts, to better understand their behaviour and capability. 

\textbf{LLMs evaluation benchmarks.~}
Understanding adherence of LLMs to factual knowledge has gained increasing attention in recent years given the widespread use of these models. Hallucination detection and mitigation has been identified as a fundamental step to ensure transparency and trustworthiness of the models. In recent years a proliferation of benchmarks for evaluating factuality of LLMs has been observed with \citep{huang2023survey} presenting a survey of existing hallucination detection and mitigation approaches. They include many established benchmarks such as TruthfulQA \citep{lin2022truthfulqa}, FreshQA \citep{vu2023freshllms}, HaluEval \citep{li2023halueval}, HalluQA \citep{cheng2023evaluating}, and FELM \citep{chen2023felm} that mainly focus on short-form answer evaluation where the knowledge of the LLM is tested in the form of a single factoid evaluated binary as true or false in adherence to the specific benchmark. More recent works \citep{wei2024longform,min-etal-2023-factscore} cover long-form answer evaluation where the answer is decomposed into individual facts that are then independently evaluated. An ensemble metric is computed at the end to represent the overall evaluation score.

These previous works include the primary definition of truthfulness and factuality as a binary concept where the goal is to test the knowledge of the LLM in the form of a single (short-form) or multiple (long-form) factoids evaluated as \textbf{true or false} given the specific benchmark. 

In this work we move away from a dualistic vision of the truth and we focus on cases where the answer to a question is not unique. We investigate how LLMs deal with real-world conflicting information where there exist different sources and possible answers considered equally trustworthy.

\section{WikiContradict}
\label{sec:dataset}
In this section, we describe the process of leveraging contradiction tags from Wikipedia to develop \texttt{WikiContradict}, a QA-based benchmark consisting of 253 human-annotated instances that cover different types of real-world knowledge conflicts.

\begin{figure}
    \centering 
    \includegraphics[width=.9\textwidth]{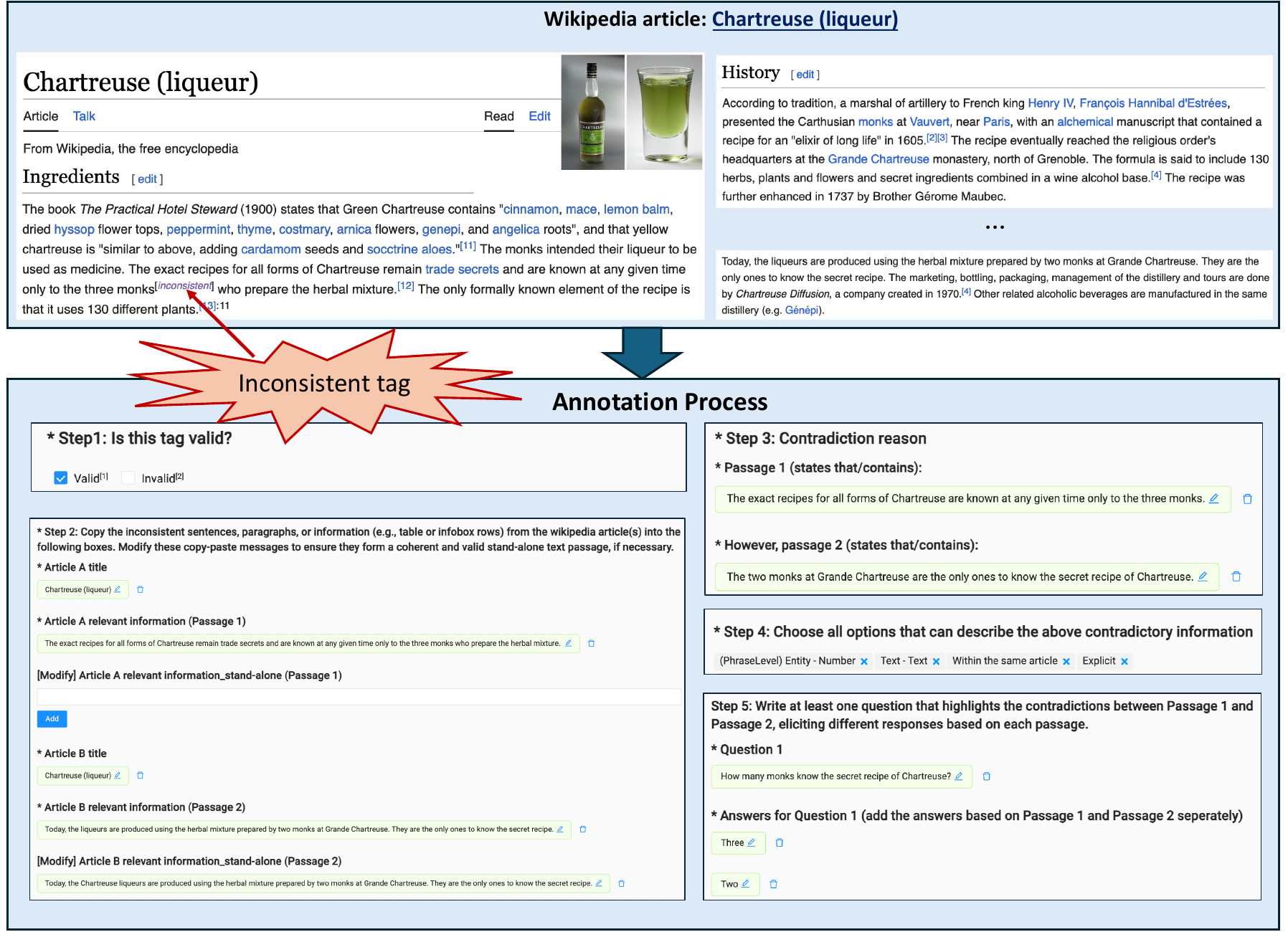}
    \caption{\texttt{WikiContradict} data annotation pipeline.}
    \label{fig:annotation}
\end{figure}

\subsection{Data collection and processing}
Although Wikipedia is widely regarded as a high-quality pre-training dataset for most LLMs, its content is not without flaws, including speculation, inconsistencies, and other content issues. To address these problems, Wikipedia editors use a wide range of maintenance tags to flag  problematic content for improvement. However, these maintenance tags are typically removed when creating Wikipedia datasets for LLM pre-training, which results in content with various quality issues being included in the pre-training process.

In this work, we focus on three tags that indicate content inconsistencies: \emph{inconsistent}, \emph{self-contradictory}, and \emph{contradict-other}. The first two tags denote contradictory statements within the same article, whereas the third tag highlights instances where the content of one article contradicts that of another article. In total, we collect around 1,200 articles that contain these tags through the Wikipedia maintenance category ``\emph{Wikipedia articles with content issues}''\footnote{\url{https://en.wikipedia.org/wiki/Category:Wikipedia_articles_with_content_issues}. The dataset is collected on Feburary 26, 2024.}. The upper portion of Figure \ref{fig:annotation} illustrates a Wikipedia article that has been flagged by an ``\emph{inconsistent}'' tag with the comment ``\emph{contradictory number of monks}''. The tagged paragraphs from these articles, together with the editors' comments whenever available, serve as the starting point for our annotations.

\subsection{Data annotation}
We developed an annotation interface using Label Studio\footnote{\url{https://labelstud.io/}}, as shown in the lower portion of Figure \ref{fig:annotation}. Given a content inconsistency tag provided by Wikipedia editors, our annotators first need to verify whether the tag is valid by checking the relevant article content, the editor's comment, as well as the information in the edit history and the article's talk page if necessary.   

For the verified valid tags, we instruct annotators to extract the two contradictory paragraphs/sentences from the original article(s), slightly modifying them as needed to create stand-alone passages. Such modifications normally require the annotators to resolve anaphoric references (e.g., \emph{She}) in the first sentence of each passage. Next, the annotators should provide a brief explanatory text to clarify the contradictions between the two passages and categorize them using a pre-defined taxonomy, such as \emph{Date, Number, Explicit, Implicit}. Implicit contradiction requires us do to some reasoning to understand why the two passages are contradicted 
whereas explicit contradiction is more straightforward, with the inconsistency clearly evident on the surface.

Finally, the annotators must craft at least one question that effectively highlights the contradictions between the two passages, eliciting different answers depending on which passage is referenced. The two examples from Figure \ref{fig:example} illustrate our annotation results. On average, annotators spent around 30 minutes to annotate a tag; longer times are required to annotate tags related to implicit conflicts, especially for cases in which the reasons of inconsistency are not explicitly mentioned in the comments from Wikipedia editors.  More details about the pre-defined contradiction taxonomy and the full annotation guideline can be found in Appendix \ref{appx:guidelines} of the supplementary materials.

\subsection{Data statistics}

\begin{wraptable}{r}{6cm}
 \vspace*{-2.5\baselineskip}
   \caption{\texttt{WikiContradict} dataset.}
   \vspace{5pt}
  \label{tab:datastat}
  \centering
  \begin{small}

  \begin{tabular}{l|l}
    \toprule
    Wikipedia tags     &     261  \\ 
    Verified valid tags     &    130  \\  
    \midrule
    Annotated instances & 253  \\ 
    \midrule
    \multicolumn{2}{c}{Question type} \\ 
        \midrule
         Yes/No questions & 133    (53\%)  \\
         Wh-questions       & 120  (47\%)\\
   \midrule
    \multicolumn{2}{c}{Contradiction type} \\ 
        \midrule
         Explicit & 161    (64\%)  \\ 
         Implicit  & 92    (36\%)\\    
         \bottomrule
  \end{tabular}
  \end{small}
\end{wraptable}

Using the annotation pipeline outlined in the previous section, the authors annotated the collected Wikipedia articles marked with inconsistency tags, yielding a total of 253 annotated instances. Each instance comprises a question, two contradictory passages, and two distinct answers, each derived from one of the passages. 
Table \ref{tab:datastat} shows an overview of the dataset statistics. Notably, among all annotated instances, approximately 61\% of questions are categorized as wh-questions seeking specific information. Furthermore, a significant proportion of instances (36\%) contain implicit contradictions.

\begin{figure}[t]
    \centering 
    \includegraphics[width=0.9\textwidth]{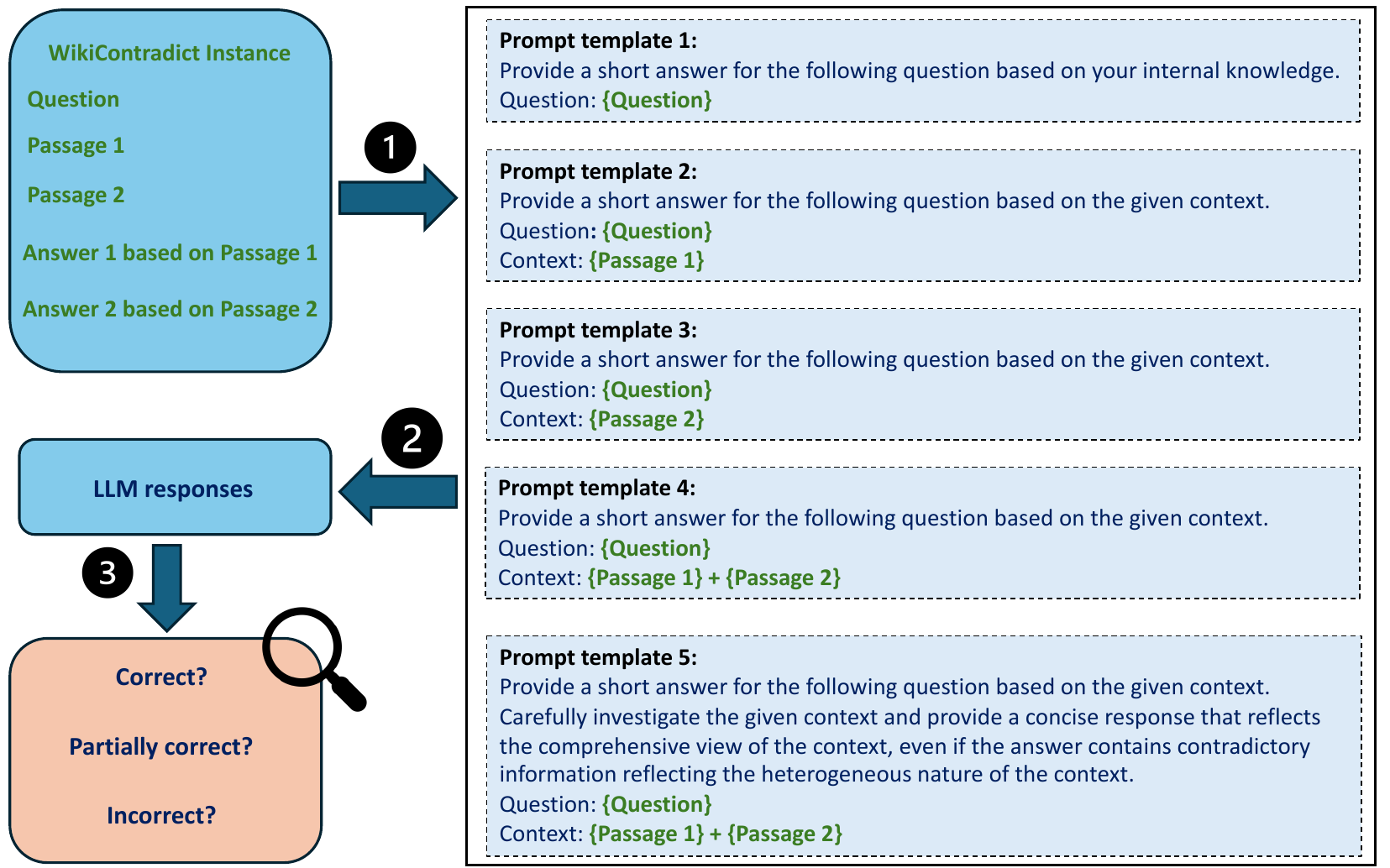}
    \caption{\texttt{WikiContradict} evaluation.}
    \label{fig:evaluation}
\end{figure}

\subsection{Evaluation}
\label{sec:evalProtocal}
To investigate how LLMs respond to real-world inter-context conflicts, we develop five prompt templates to evaluate their performance under different question-answering (QA) scenarios.  As illustrated in Figure \ref{fig:evaluation}, for each annotated instance from the \texttt{WikiContradict} dataset, we generate five question prompts based on these pre-defined templates. Specifically, template 1 evaluates a model's internal knowledge, while templates 2 and 3 examine its performance in the RAG setting with a single retrieved passage. Templates 4 and 5, on the other hand, assess a model's ability to handle QA in the RAG setting with two contradictory passages that can lead to different answers.

To evaluate LLMs' responses to these question prompts, we follow the relaxed evaluation mode from FreshLLM \citep{vu2023freshllms} by allowing additional hallucinated or inaccurate information as long as the primary answer is accurate and any additional information does not contradict with the primary answer. More specifically, each response is evaluated as ``\emph{correct}'', ``\emph{partially correct}'', or ``\emph{incorrect}'':
\begin{itemize}
    \item ``\emph{Correct}'' if the response accurately matches all the answers in the annotated answer list. For prompt templates 4 and 5, the response should identify and contain the contradictory answers that reflect the heterogeneous nature of the context. Additionally, the correct response should not indicate a preference for one answer over another, and it should not combine two different correct answers without indicating the contradictory nature of these answers.
    \item ``\emph{Partially correct}'' applies to prompt templates 1, 4, and 5; it means that the response only matches one of the answers in the annotated answer list, or the response matches all the answers in the correct answer list but indicates a preference for one answer over another.
    \item ``\emph{Incorrect}'' if the response does not match any of the annotated answers, or the response merely combines two contradictory answers from the annotated answer list and indicates that both are possible at the same time without indicating the contradictory nature of the two context passages.
\end{itemize}
  
Following the criteria described above, for the question from Example 2 of Figure \ref{fig:example}, an LLM response to prompt template 4, ``\ul{\emph{According to the context, the number of monks who know the secret recipe of Chartreuse is either three or two. The first statement suggests that three monks know the recipe, while the second statement contradicts this, stating that only two monks at Grande Chartreuse know the secret recipe.}}'' is a \emph{\textbf{correct}} response. In contrast, the LLM response to prompt template 3, ``\ul{\emph{Two monks know the secret recipe of Chartreuse.}}'' is \emph{\textbf{partially correct}}, as it only captures one aspect of the contradictory information presented in the context.

\section{Human evaluation: LLMs are struggling on WikiContradict}
\label{sec:human_eval}

To understand LLMs' behavior when faced with real-world inter-context conflicts, we use \texttt{WikiContradict} to benchmark a list of LLMs with the evaluation protocol described in Section \ref {sec:evalProtocal}\footnote{In our experiments, decoding temperature=0, maximum output length=250 tokens}.
The authors independently evaluated a subset of answers, comprising 1,375 responses from 5 LLMs based on the five prompt templates, as shown in Figure \ref{fig:evaluation}, for 55 instances. Each response is assessed by two authors of this paper, yielding a total of 2,750 human judgements. The inter-annotator agreements, as measured by Cohen's kappa $\kappa$, were moderate to substantial, with values of 0.58 for prompt template 3, 0.67 for prompt template 4, 0.84 for prompt template 0, 0.85 for prompt template 2, and 0.88 for prompt template 1. After resolving the annotation disagreements among annotators, our final human evaluation study dataset (\texttt{WikiContradict\_HuamEval}) consists of 1,200 samples resulting from five LLMs’ responses to 48 WikiContradict instances based on five prompt templates\footnote{We exclude 7 instances from our human evaluation study due to ambiguity in their inconsistencies, which are not crystal clear. These instances are also excluded from the final \texttt{WikiContradict} dataset.}.

\begin{table}[t]
   \caption{Human evaluation results on \texttt{WikiContradict\_HumanEval}. ``C'', ``PC'' and ``IC'' stand for ``\emph{Correct}'', ``\emph{Partially correct}'', ``\emph{Incorrect}'', respectively. ``all'', ``exp'', and ``imp'' represent for instance types: all instances, instances with explicit conflicts, and instances with implicit conflicts. The numbers represent the ratio of responses from each LLM that were assessed as ``\emph{Correct}, ``\emph{Partially correct}, or ``\emph{Incorrect} for each instance type under a prompt template. The bold numbers highlight the best models that correctly answer questions for  each type and prompt template.}
   \label{tab:humeval}
    \begin{footnotesize}
    \begin{tabular}{|p{0.25cm}|p{0.46cm}|p{0.475cm}|p{0.46cm}|p{0.46cm}|p{0.475cm}|p{0.46cm}|p{0.46cm}|p{0.46cm}|p{0.46cm}|p{0.46cm}|p{0.475cm}|p{0.46cm}|p{0.46cm}|p{0.475cm}|p{0.46cm}|}
   \hline
        & \multicolumn{3}{c|}{Mistral-7b-inst} & \multicolumn{3}{c|}{Mixtral-8x7b-inst} & \multicolumn{3}{c|}{Llama-2-70b-chat}& \multicolumn{3}{c|}{Llama-3-70b-inst} & \multicolumn{3}{c|}{GPT-4} \\ \hline
        ~ & \textbf{all}  & \textbf{exp}& \textbf{imp} & \textbf{all} & \textbf{exp} & \textbf{imp} & \textbf{all}  & \textbf{exp} & \textbf{imp}& \textbf{all} & \textbf{exp} & \textbf{imp} & \textbf{all}  & \textbf{exp} & \textbf{imp} \\ \hline
        \multicolumn{16}{|c|}{\textbf{Prompt Template 1}} \\ \hline
        C & \textbf{4.2} & \textbf{6.7} & 0.0 & 2.1 & 0.0 & 5.9 & 0.0 & 0.0 & 0.0 & \textbf{4.2} & 0.0 & \textbf{11.8} & 2.1 & 0.0 & 5.9 \\ \hline
        PC & 33.3 & 23.3 & 47.1 & 52.1 & 43.3 & 64.7 & 54.2 & 43.3 & 70.6 & 52.1 & 46.7 & 58.8 & 58.3 & 53.3 & 64.7 \\ \hline
        IC & 62.5 & 70.0 & 52.9 & 45.8 & 56.7 & 29.4 & 45.8 & 56.7 & 29.4 & 43.8 & 53.3 & 29.4 & 39.6 & 46.7 & 29.4 \\ \hline
        \multicolumn{16}{|c|}{\textbf{Prompt Template 2}}\\ \hline
        C & 92.7 & - & - & \textbf{97.6} & - & - & 87.8 & - & - & 95.1 & - & - & \textbf{97.6} & - & - \\ \hline
        IC & 7.3 & - & - & 2.4 & - & - & 12.2 & - & - & 4.9 & - & - & 2.4 & - & - \\ \hline
        \multicolumn{16}{|c|}{\textbf{Prompt Template 3}}\\ \hline
        C & 82.9 & -& - & \textbf{92.7} & - & - & 90.2 & - & - & \textbf{92.7} & - & - & 87.8 & - & - \\ \hline
        IC & 17.1 & - & - & 7.3 & - & - & 9.8 & - & - & 7.3 & - & - & 12.2 & - & - \\ \hline
         \multicolumn{16}{|c|}{\textbf{Prompt Template 4}} \\ \hline
       C & 2.1 & 3.3 & 0.0 & 4.2 & 3.3 & 5.9 & 4.2 & 3.3 & 5.9 & \textbf{10.4} & \textbf{13.3} & 5.9 & 6.3 & 3.3 & \textbf{11.8} \\ \hline
        PC & 87.5 & 86.7 & 88.2 & 91.7 & 93.3 & 88.2 & 93.8 & 96.7 & 88.2 & 81.3 & 80.0 & 82.4 & 85.4 & 96.7 & 64.7 \\ \hline
        IC & 10.4 & 10.0 & 11.8 & 4.2 & 3.3 & 5.9 & 2.1 & 0.0 & 5.9 & 8.3 & 6.7 & 11.8 & 8.3 & 0.0 & 23.5 \\ \hline
        \multicolumn{16}{|c|}{\textbf{Prompt Template 5}} \\ \hline
        C & 20.8 & 26.7 & 11.8 & 14.6 & 16.7 & 11.8 & 22.9 & 26.7 & \textbf{17.6} & \textbf{43.8} & \textbf{60.0} & \textbf{17.6} & 10.4 & 10.0 & 11.8 \\ \hline
        PC & 70.8 & 63.3 & 82.4 & 83.3 & 83.3 & 82.4 & 68.8 & 63.3 & 76.5 & 45.8 & 26.7 & 76.5 & 81.3 & 90.0 & 64.7 \\ \hline
        IC & 8.3 & 10.0 & 5.9 & 2.1 & 0.0 & 5.9 & 8.3 & 10.0 & 5.9 & 10.4 & 13.3 & 5.9 & 8.3 & 0.0 & 23.5 \\ \hline
        \end{tabular}
    \end{footnotesize}
\end{table}

Table \ref{tab:humeval} presents the results of \texttt{WikiContradict\_HumanEval} for 5 LLMs: \emph{Mistral-7b-instruct}, \emph{Mixtral-8x7b-instruct}, \emph{Llama2-2-70b-chat}, \emph{Llama3-70b-instruct}, and \emph{GPT-4-turbo-2024-04-09}. The table provides a detailed breakdown of response accuracy for each LLM, categorized into three types: \emph{correct}, \emph{partially correct} (applicable to prompt templates 1, 4, and 5), and \emph{incorrect}. We further distinguish between instances with explicit conflicts (30 instances) and implicit conflicts (18 instances) to provide a more nuanced understanding of the LLMs' performance. Below we summarise a few key observations on \texttt{WikiContradict\_HumanEval}.

\textbf{Prompt template 1: There is a significant overlap between the internal knowledge of LLMs and the content of Wikipedia.} As expected, the portion of correct or partially correct answers based on their internal knowledge, ranges from 37.5\% (\emph{Mistral-7b-inst}) to 60.4\% (\emph{GPT-4}). 

\textbf{Prompt template 2 and 3: When tasked with answering questions based on a provided context, LLMs are generally capable of generating correct responses for the majority of instances, as long as the context does not contain conflicting information.} However, we observe that \emph{GPT-4} exhibits a ``stubborn'' behavior, particularly with prompt template 3. It often relies on its internal knowledge, which may not align with the given context, resulting in a lower accuracy of 87.8\% compared to \emph{Mixtral-8X7b-inst} and \emph{Llama3-3-70b-inst}, which perform better in this scenario.

\textbf{Prompts template 4 and 5: LLMs often struggle to provide correct answers when the context contains conflicting information.} Typically, the models rely on a single passage to inform their response, disregarding the other passage. Notably, some models attempt to reconcile the conflicting information by providing both answers, but then proceed to explain why one of them is incorrect. This phenomenon is particularly pronounced in the \emph{Mistral-7b-inst} model.

\textbf{Prompts template 4 vs. 5: LLMs can improve their performance in providing correct answers when explicitly instructed to consider conflicting information within the given context.} Notably, Llama-3-70b-inst exhibits the most substantial improvement, jumping from 10.4\% to 43.8\%. In contrast, GPT-4 demonstrates the smallest improvement, increasing from 6.3\% to 10.4\%, which is likely attributed to its previously observed stubborn behavior in prompt template 3.

\textbf{Explicit conflicts vs. implicit conflicts: When explicitly instructed to consider conflicting information within the given context, LLMs' performance in providing correct answers improves in particular 
in cases where conflicts are explicitly stated.} For instance, for \emph{Llama-3-70b-inst}, the performance on explicit conflicts instances jumps from 13.3\% to 60.0\%, while the performance on implicit conflicts instances improves from 5.9\% to 17.6\%.

\textbf{Additional evaluations: }We conducted additional human evaluation studies on 48 instances from \texttt{WikiContradict\_HumanEval} using four variations of prompt template 5: prompt template 5.1 swaps the positions of passage 1 and passage 2 from the original template 5; prompt template 5.2 instructs LLMs to identify any contradictions in the given context with respect to the question; prompt template 5.3 provides LLMs with manually revised passage 1 and passage 2, where contradictions with respect to the question were resolved; prompt template 5.4 tasks LLMs with detecting any contradictions in the revised consistent context from template 5.3 with respect to the question. In this experiment, each response is assessed by a single human annotator.

The human evaluation of 5 LLMs on 4 prompt templates (5, 5.1, 5.2, and 5.3) reveals the following insights. 
1) \textbf{No position bias}: the results for prompt templates 5 and 5.1 are similar for all LLMs;
2) \textbf{Contradiction detection}: for prompt template 5.2, all LLMs perform better in detecting contradictions in the given context compared to generating correct answers for prompt template 5. \emph{GPT-4} and \emph{Llama-3-70b-instruct} are the top performers, with \emph{GPT-4} detecting contradictions and providing reasonable explanations in around 88\% of cases, followed by \emph{Llama-3-70b-instruct} with 77\%;
3) \textbf{Conflict-free context}: for prompt template 5.3, all models demonstrate high performance in correctly answering questions based on the given context (with correct response rates ranging from 85\% to 87\%), where there are no conflicts between passage 1 and passage 2. This is consistent with the observations for prompt templates 2 and 3 in the previous section;
4) \textbf{Difficulty in detecting conflict-free contexts}: for prompt template 5.4, all models struggle to detect when there are no contradictions in the given context, performing worse than when generating correct answers for prompt template 5.3. \emph{GPT-4} and \emph{Llama-3-70b-instruct} are the worst performers, only confirming 44\% and 48\% of instances as conflict-free, and ``making up'' reasons to explain why the two passages conflict in the remaining instances. See Appendix \ref{appx:details_prompt} in the supplementary materials for more on the prompt templates 5.1 - 5.4 and human evaluation results.

\section{Automatic evaluation: WikiContradictEval}
\label{sec:auto_eval}

Since human evaluation is costly and time-consuming, to facilitate future evaluations, we have developed \texttt{WikiContradictEval}, a simple automatic evaluation method that leverages few-shot in-context learning to teach LLMs to judge model responses for prompt template 5, which is aligned with the central focus of \texttt{WikiContradict} benchmark to evaluate LLMs on real-world inter-context conflicts. Table \ref{tab:auto_eval_llmjudge} reports the results of different judge LLMs on the testing dataset that consists of 240 responses to prompt template 5 from five testing LLMs in \texttt{WikiContradict\_HumanEval}. Among the four judge LLMs evaluated, the top-performing model, \emph{GPT-4}, achieves an F-score of 82.5 in accurately identifying correct responses, with a precision score of 73.4 and a recall score of 94.0. The best open-source model, \emph{Llama-3-70b-inst}, closely follows, with an overall F-score of 80.4 in identifying correct responses. For further details on the judge LLM prompt, please refer to Appendix \ref{appx:details_judge} in the supplementary materials.

\begin{table}[t]
    \centering
    \caption{Judge LLM results for prompt template 5 on \texttt{WikiContradict\_HumanEval}. GTP-4 and GPT-4o represent ``gpt-4-turbo-2024-04-09'' and ``gpt-4o-2024-05-13'', respectively.}
    \label{tab:auto_eval_llmjudge}
    \begin{small}
    \begin{tabular}{|l|c|c|c|c|c|}
    \hline

        ~ &  &  & \emph{\textbf{Correct}}  &\emph{\textbf{Partially correct}}  & \emph{\textbf{Incorrect}}   \\ 
        ~ & \textbf{Acc} & \textbf{Macro-F} & \textbf{(P/R/F)} & \textbf{(P/R/F)} &  \textbf{(P/R/F)}  \\ \hline    
  Mixtral-8x7b-inst & 26.2 & 19.7 &  10.5 / 22.0 / 14.2  &  66.2 / 30.0 / 41.3  &  2.9 / 5.0 / 3.7 \\ \hline
        Llama-3-70b-inst & 85.4 & 74.4 &  72.6 / 90.0 / 80.4  &  96.1 / 87.6 / 91.7  &  47.8 / 55.0 / 51.2 \\ \hline
        GPT-4 & \textbf{86.7} & \textbf{76.1} &  \textbf{73.4 / 94.0 / 82.5}  &  \textbf{96.8 / 88.2 / 92.3}  &  \textbf{52.4 / 55.0 / 53.7} \\ \hline
        GPT-4o& 83.3 & 71.3 &  74.5 / 76.0 / 75.2  &  94.9 / 88.2 / 91.5  &  38.7 / 60.0 / 47.1 \\ \hline
    \end{tabular}
    \end{small}
\end{table}

\begin{table}[t]
    \centering
    \caption{Comparing human judgement with Llama3-70b-instruct judgement for prompt template 5 for five testing LLMs on \texttt{WikiContradict\_HumanEval}. }
    \label{tab:auto_eval_comp}
    \begin{small}
    \begin{tabular}{|l|c|c|c|c|c|c|}
    \hline
        & \multicolumn{3}{c|}{\textbf{Human judgement}}& \multicolumn{3}{c|}{\textbf{LLM judgement}} \\ \hline
        ~ & \emph{Correct} & \emph{Partially correct} & \emph{Incorrect} & \emph{Correct} & \emph{Partially correct} & \emph{Incorrect} \\ \hline
        Mistral-7b-inst & 20.8 & 70.8 & 8.3 & 39.6 & 54.2 & 6.2 \\ \hline
        Mixtral-8x7b-inst & 14.6 & 83.3 & 2.1 & 16.7 & 77.1 & 6.2 \\ \hline
        Llama-2-70b-chat & 22.9 & 68.8 & 8.3 & 22.9 & 64.6 & 12.5 \\ \hline
        Llama-3-70b-inst & 43.8 & 45.8 & 10.4 & 43.8 & 41.7 & 14.6 \\ \hline
        GPT-4 & 10.4 & 81.3 & 8.3 & 8.3 & 81.2 & 10.4 \\ \hline
    \end{tabular}
    \end{small}
\end{table}

Table \ref{tab:auto_eval_comp} presents a comparison of human judgments with the assessments generated by the \emph{Llama-3-70b-inst} judge model for prompt template 5 on the \texttt{WikiContradict\_HumanEval} dataset. Note that our LLM judge model is generally well-aligned with human judgment in identifying correct responses for most LLMs, with one notable exception. The \emph{Mistral-7b-instruct} model often attempts to reconcile conflicting information by providing both answers, which our LLM judge model mistakenly views as valid responses, as a result, it overestimates the correct responses.

Finally we apply the best open-source judge model, \emph{Llama-3-70b}, to assess a list of seven LLMs based on prompt template 5 on all 253 instances from the \texttt{WikiContradict} dataset. Table \ref{tab:auto_eval_all} shows the evaluated results for each LLM. Among all testing models, \emph{Mistral-7b-inst} and \emph{Llama-3-70b-inst} are the top performers in correctly answering questions, with correct response rates of 50.6\% and 45.8\%, respectively. It is noteworthy that, based on the above analysis, there is a high probability that the judge LLM overestimates the performance of \emph{Mistral-7b-inst}, suggesting a potential bias in its evaluation. In addition, all models except Flan-ul2 have higher correct response rates for instances with explicit contradictions compared to instances with implicit contradictions, which is aligned with the observation in our human evaluation study (Section \ref{sec:human_eval}).

\begin{table}[h]
    \centering

    \caption{\emph{Llama-3-70b} judge results on \texttt{WikiContradict} based on prompt template 5. ``C'', ``PC'', and ``IC'' stand for ``\emph{Correct}'', ``\emph{Partially correct}'', ``\emph{Incorrect}'', respectively.}
    \label{tab:auto_eval_all}
    \begin{small}
    \begin{tabular}{|l|p{0.7cm}|p{0.7cm}|p{0.7cm}|p{0.8cm}|p{0.8cm}|p{0.8cm}|p{0.8cm}|p{0.8cm}|p{0.8cm}|}
    \hline
        & \multicolumn{3}{c|}{\textbf{All instances}}& \multicolumn{3}{c|}{\textbf{Explicit contradictions }} & \multicolumn{3}{c|}{\textbf{Implicit contradictions}}\\ \hline
        ~ & \emph{C} & \emph{PC} & \emph{IC} & \emph{C} & \emph{PC} & \emph{IC} & \emph{C} & \emph{PC} & \emph{IC}\\ \hline
        Mistral-7b-inst& 50.6 & 39.9 & 5.9 & 57.8 & 36 & 3.1 & 38.0 & 46.7 & 10.9 \\ \hline
        Llama-3-70b-inst & 45.8 & 38.7 & 13.8 & 55.9 & 28.6 & 14.3 & 28.3 & 56.5 & 13.0 \\ \hline
        Llama-3-8b-inst & 37.9 & 53 & 6.7 & 47.8 & 46.6 & 5.0 & 20.7 & 64.1 & 9.8 \\ \hline
        Mixtral-8x7b-inst & 37.9 & 52.2 & 6.3 & 43.5 & 48.4 & 3.7 & 28.3 & 58.7 & 10.9 \\ \hline
        GPT-4 & 15.0 & 73.5 & 11.5 & 19.3 & 70.8 & 9.9 & 7.6 & 78.3 & 14.1 \\ \hline
        Llama-2-13b-chat & 11.5 & 73.5 & 9.5 & 12.4 & 74.5 & 7.5 & 9.8 & 71.7 & 13.0 \\ \hline
        Flan-ul2 & 1.2 & 90.1 & 7.9 & 0 & 92.5 & 6.8 & 3.3 & 85.9 & 9.8 \\ \hline    \end{tabular}
     \end{small}
\end{table}

\section{Limitation and future work} 
\label{sec:limit}
We note several limitations of the current benchmark. First, the benchmark presented in this paper is limited to English language instances. Given contradicting statements is quite a nuanced challenge any solution evaluated using this benchmark may not generalise to other languages. Second, the reliance on Wikipedia tags may skew the benchmark towards specific types of contradictory statements. Third, the contradictions covered in this benchmark are limited to text-text contradictions. Contradictions can occur across modalities, for example instances of annotation tag have been used to capture statements in the text that describe a phenomena that differs from what is shown in an image. Future work will aim to include coverage across different modalities. 

\section{Conclusion}
\label{sec:conc}

Unlike most previous work on LLM knowledge conflicts within RAG frameworks, which focus on ``\emph{context-memory conflicts}'', our focus is on ``\emph{real-world inter-context conflicts}''. Within this setting we introduced the \texttt{WikiContradict} benchmark, which consists of 253 human-annotated instances covering different types of contradictions identified by Wikipedia editors. Our annotation of these instances, which includes isolation of relevant conflicting passages, resolution of anaphora and the creation of at least one question that highlights the contradictions, results in a benchmark that can effectively evaluate the capacity of LLMs to manage and reason over knowledge conflicts. This capacity was highlighted by the results of the evaluation of LLMs on the benchmark, in which for each contradictory instance, the LLMs were given different prompts, each one containing a different instruction and/or different conflicting information contained within the instance. Our experiments show that LLMs often struggle to correctly identify and manage real-world inter-context conflicts, indicating the usefulness of the benchmark and the need for further research in this direction.

\bibliography{lib}

\begin{thebibliography}{18}
\providecommand{\natexlab}[1]{#1}
\providecommand{\url}[1]{\texttt{#1}}
\expandafter\ifx\csname urlstyle\endcsname\relax
  \providecommand{\doi}[1]{doi: #1}\else
  \providecommand{\doi}{doi: \begingroup \urlstyle{rm}\Url}\fi

\bibitem[Brown et~al.(2020)Brown, Mann, Ryder, Subbiah, Kaplan, Dhariwal, Neelakantan, Shyam, Sastry, Askell, Agarwal, Herbert-Voss, Krueger, Henighan, Child, Ramesh, Ziegler, Wu, Winter, Hesse, Chen, Sigler, Litwin, Gray, Chess, Clark, Berner, McCandlish, Radford, Sutskever, and Amodei]{LLMFewShortLearner}
T.~Brown, B.~Mann, N.~Ryder, M.~Subbiah, J.~D. Kaplan, P.~Dhariwal, A.~Neelakantan, P.~Shyam, G.~Sastry, A.~Askell, S.~Agarwal, A.~Herbert-Voss, G.~Krueger, T.~Henighan, R.~Child, A.~Ramesh, D.~Ziegler, J.~Wu, C.~Winter, C.~Hesse, M.~Chen, E.~Sigler, M.~Litwin, S.~Gray, B.~Chess, J.~Clark, C.~Berner, S.~McCandlish, A.~Radford, I.~Sutskever, and D.~Amodei.
\newblock Language models are few-shot learners.
\newblock In H.~Larochelle, M.~Ranzato, R.~Hadsell, M.~Balcan, and H.~Lin, editors, \emph{Advances in Neural Information Processing Systems}, volume~33, pages 1877--1901. Curran Associates, Inc., 2020.
\newblock URL \url{https://proceedings.neurips.cc/paper_files/paper/2020/file/1457c0d6bfcb4967418bfb8ac142f64a-Paper.pdf}.

\bibitem[Chen et~al.(2023)Chen, Zhao, Zhang, Chern, Gao, Liu, and He]{chen2023felm}
S.~Chen, Y.~Zhao, J.~Zhang, I.-C. Chern, S.~Gao, P.~Liu, and J.~He.
\newblock Felm: Benchmarking factuality evaluation of large language models.
\newblock In \emph{Thirty-seventh Conference on Neural Information Processing Systems Datasets and Benchmarks Track}, 2023.
\newblock URL \url{http://arxiv.org/abs/2310.00741}.

\bibitem[Cheng et~al.(2023)Cheng, Sun, Zhang, Wang, Liu, Zhang, He, Huang, Yin, Chen, and Qiu]{cheng2023evaluating}
Q.~Cheng, T.~Sun, W.~Zhang, S.~Wang, X.~Liu, M.~Zhang, J.~He, M.~Huang, Z.~Yin, K.~Chen, and X.~Qiu.
\newblock Evaluating hallucinations in chinese large language models, 2023.

\bibitem[Gebru et~al.(2021)Gebru, Morgenstern, Vecchione, Vaughan, Wallach, Iii, and Crawford]{gebru2021datasheets}
T.~Gebru, J.~Morgenstern, B.~Vecchione, J.~W. Vaughan, H.~Wallach, H.~D. Iii, and K.~Crawford.
\newblock Datasheets for datasets.
\newblock \emph{Communications of the ACM}, 64\penalty0 (12):\penalty0 86--92, 2021.

\bibitem[Huang et~al.(2023)Huang, Yu, Ma, Zhong, Feng, Wang, Chen, Peng, Feng, Qin, and Liu]{huang2023survey}
L.~Huang, W.~Yu, W.~Ma, W.~Zhong, Z.~Feng, H.~Wang, Q.~Chen, W.~Peng, X.~Feng, B.~Qin, and T.~Liu.
\newblock A survey on hallucination in large language models: Principles, taxonomy, challenges, and open questions, 2023.

\bibitem[Jin et~al.(2024)Jin, Cao, Chen, Liu, Jiang, Xu, Qiuxia, and Zhao]{jin-etal-2024-tug-war}
Z.~Jin, P.~Cao, Y.~Chen, K.~Liu, X.~Jiang, J.~Xu, L.~Qiuxia, and J.~Zhao.
\newblock Tug-of-war between knowledge: Exploring and resolving knowledge conflicts in retrieval-augmented language models.
\newblock In N.~Calzolari, M.-Y. Kan, V.~Hoste, A.~Lenci, S.~Sakti, and N.~Xue, editors, \emph{Proceedings of the 2024 Joint International Conference on Computational Linguistics, Language Resources and Evaluation (LREC-COLING 2024)}, pages 16867--16878, Torino, Italia, May 2024. ELRA and ICCL.
\newblock URL \url{https://aclanthology.org/2024.lrec-main.1466}.

\bibitem[Lewis et~al.(2020)Lewis, Perez, Piktus, Petroni, Karpukhin, Goyal, K\"{u}ttler, Lewis, Yih, Rockt\"{a}schel, Riedel, and Kiela]{rag1}
P.~Lewis, E.~Perez, A.~Piktus, F.~Petroni, V.~Karpukhin, N.~Goyal, H.~K\"{u}ttler, M.~Lewis, W.-t. Yih, T.~Rockt\"{a}schel, S.~Riedel, and D.~Kiela.
\newblock Retrieval-augmented generation for knowledge-intensive nlp tasks.
\newblock In H.~Larochelle, M.~Ranzato, R.~Hadsell, M.~Balcan, and H.~Lin, editors, \emph{Advances in Neural Information Processing Systems}, volume~33, pages 9459--9474. Curran Associates, Inc., 2020.
\newblock URL \url{https://proceedings.neurips.cc/paper_files/paper/2020/file/6b493230205f780e1bc26945df7481e5-Paper.pdf}.

\bibitem[Lewis et~al.(2021)Lewis, Perez, Piktus, Petroni, Karpukhin, Goyal, Küttler, Lewis, tau Yih, Rocktäschel, Riedel, and Kiela]{lewis2021retrievalaugmented}
P.~Lewis, E.~Perez, A.~Piktus, F.~Petroni, V.~Karpukhin, N.~Goyal, H.~Küttler, M.~Lewis, W.~tau Yih, T.~Rocktäschel, S.~Riedel, and D.~Kiela.
\newblock Retrieval-augmented generation for knowledge-intensive nlp tasks, 2021.

\bibitem[Li et~al.(2023)Li, Cheng, Zhao, Nie, and Wen]{li2023halueval}
J.~Li, X.~Cheng, W.~X. Zhao, J.-Y. Nie, and J.-R. Wen.
\newblock Halueval: A large-scale hallucination evaluation benchmark for large language models.
\newblock In \emph{Proceedings of the 2023 Conference on Empirical Methods in Natural Language Processing}, pages 6449--6464, 2023.

\bibitem[Li et~al.(2024)Li, Raheja, and Kumar]{li2024contradoc}
J.~Li, V.~Raheja, and D.~Kumar.
\newblock Contradoc: Understanding self-contradictions in documents with large language models, 2024.

\bibitem[Lin et~al.(2022)Lin, Hilton, and Evans]{lin2022truthfulqa}
S.~Lin, J.~Hilton, and O.~Evans.
\newblock Truthfulqa: Measuring how models mimic human falsehoods.
\newblock In \emph{Proceedings of the 60th Annual Meeting of the Association for Computational Linguistics (Volume 1: Long Papers)}, pages 3214--3252, 2022.
\newblock URL \url{https://aclanthology.org/2022.acl-long.229.pdf}.

\bibitem[Longpre et~al.(2021)Longpre, Perisetla, Chen, Ramesh, DuBois, and Singh]{longpre-etal-2021-entity}
S.~Longpre, K.~Perisetla, A.~Chen, N.~Ramesh, C.~DuBois, and S.~Singh.
\newblock Entity-based knowledge conflicts in question answering.
\newblock In M.-F. Moens, X.~Huang, L.~Specia, and S.~W.-t. Yih, editors, \emph{Proceedings of the 2021 Conference on Empirical Methods in Natural Language Processing}, pages 7052--7063, Online and Punta Cana, Dominican Republic, Nov. 2021. Association for Computational Linguistics.
\newblock \doi{10.18653/v1/2021.emnlp-main.565}.
\newblock URL \url{https://aclanthology.org/2021.emnlp-main.565}.

\bibitem[Min et~al.(2023)Min, Krishna, Lyu, Lewis, Yih, Koh, Iyyer, Zettlemoyer, and Hajishirzi]{min-etal-2023-factscore}
S.~Min, K.~Krishna, X.~Lyu, M.~Lewis, W.-t. Yih, P.~Koh, M.~Iyyer, L.~Zettlemoyer, and H.~Hajishirzi.
\newblock {FA}ct{S}core: Fine-grained atomic evaluation of factual precision in long form text generation.
\newblock In H.~Bouamor, J.~Pino, and K.~Bali, editors, \emph{Proceedings of the 2023 Conference on Empirical Methods in Natural Language Processing}, pages 12076--12100, Singapore, Dec. 2023. Association for Computational Linguistics.
\newblock \doi{10.18653/v1/2023.emnlp-main.741}.
\newblock URL \url{https://aclanthology.org/2023.emnlp-main.741}.

\bibitem[Vu et~al.(2023)Vu, Iyyer, Wang, Constant, Wei, Wei, Tar, Sung, Zhou, Le, and Luong]{vu2023freshllms}
T.~Vu, M.~Iyyer, X.~Wang, N.~Constant, J.~Wei, J.~Wei, C.~Tar, Y.-H. Sung, D.~Zhou, Q.~Le, and T.~Luong.
\newblock Freshllms: Refreshing large language models with search engine augmentation, 2023.

\bibitem[Wang et~al.(2023)Wang, Feng, Wang, Shi, Balachandran, He, and Tsvetkov]{wang2023resolving}
Y.~Wang, S.~Feng, H.~Wang, W.~Shi, V.~Balachandran, T.~He, and Y.~Tsvetkov.
\newblock Resolving knowledge conflicts in large language models, 2023.

\bibitem[Wei et~al.(2024)Wei, Yang, Song, Lu, Hu, Huang, Tran, Peng, Liu, Huang, Du, and Le]{wei2024longform}
J.~Wei, C.~Yang, X.~Song, Y.~Lu, N.~Hu, J.~Huang, D.~Tran, D.~Peng, R.~Liu, D.~Huang, C.~Du, and Q.~V. Le.
\newblock Long-form factuality in large language models, 2024.

\bibitem[Xie et~al.(2024)Xie, Zhang, Chen, Lou, and Su]{xie2024knowledgeconflict}
J.~Xie, K.~Zhang, J.~Chen, R.~Lou, and Y.~Su.
\newblock Adaptive chameleon or stubborn sloth: Revealing the behavior of large language models in knowledge conflicts.
\newblock In \emph{The Twelfth International Conference on Learning Representations}, 2024.
\newblock URL \url{https://openreview.net/forum?id=auKAUJZMO6}.

\bibitem[Xu et~al.(2024)Xu, Qi, Wang, Wang, Zhang, and Xu]{xu2024knowledge}
R.~Xu, Z.~Qi, C.~Wang, H.~Wang, Y.~Zhang, and W.~Xu.
\newblock Knowledge conflicts for llms: A survey, 2024.

\end{thebibliography}
\bibliographystyle{abbrvnat}

\newpage

\appendix

\section{Annotation Guidelines}

\label{appx:guidelines}

Here we include the details about the pre-defined contradiction taxonomy and the full annotation guideline. The annotation interface was developed using Label Studio\footnote{\url{https://labelstud.io/}}.

\includepdf[pages=-]{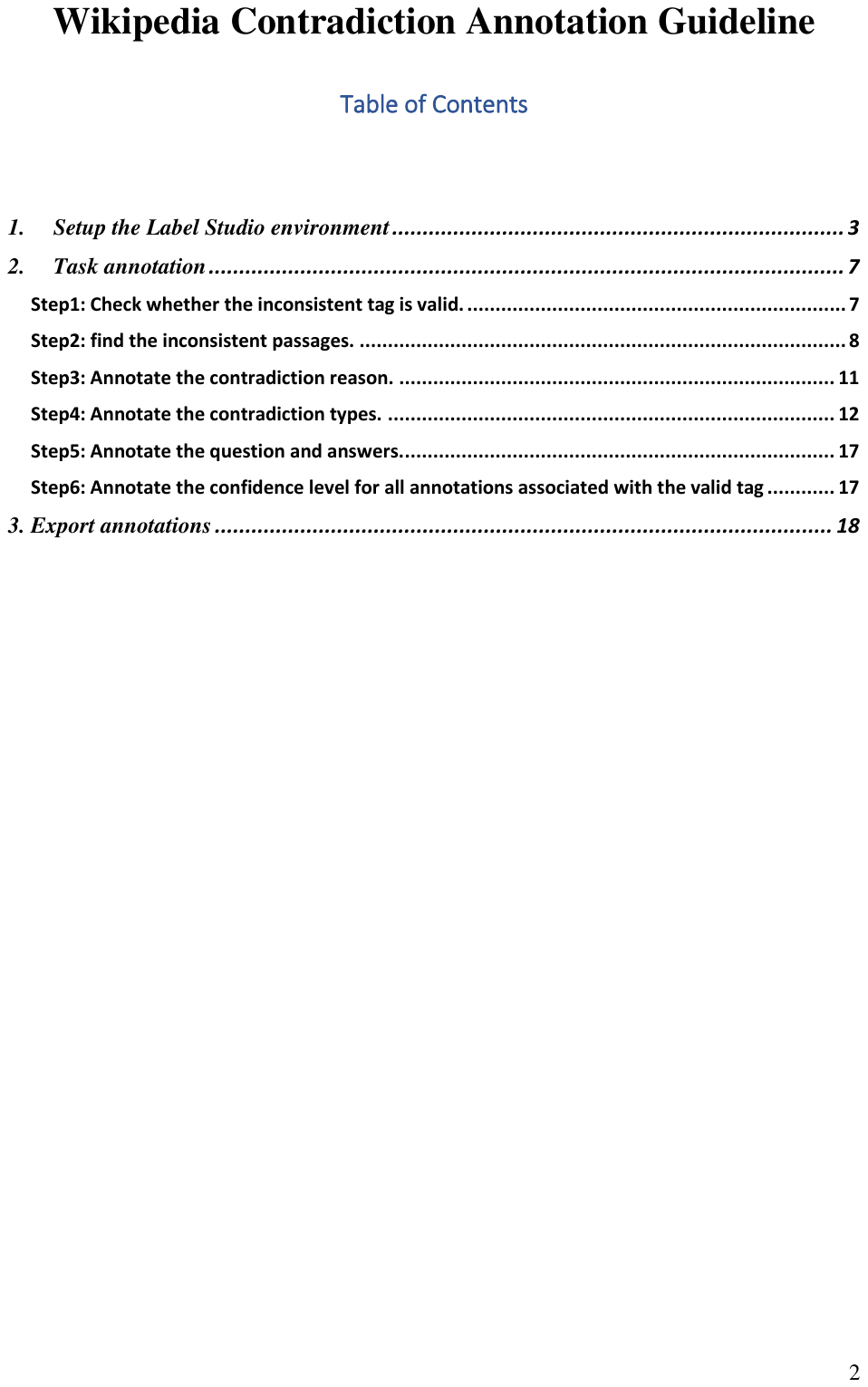}

\section{Details on the Prompt Templates 5.1 - 5.4 and Human Evaluation Results}

\label{appx:details_prompt}

Here we include the details of the Prompt Templates 5.1 - 5.4 and Human Evaluation Results (see Section 4 of the paper).

\includepdf[pages=-]{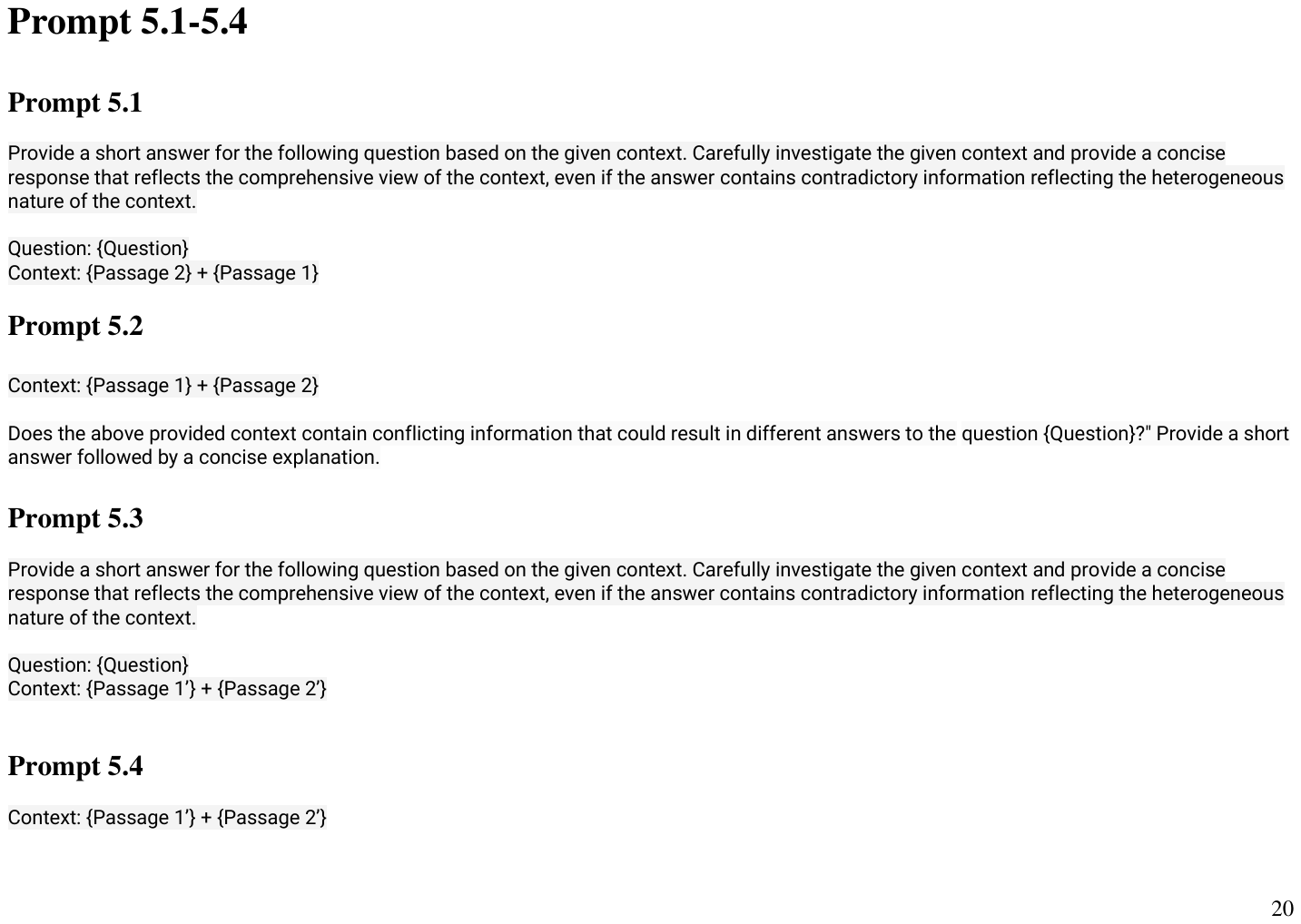}

\section{Further Details on the Judge LLM Prompt}

\label{appx:details_judge}

Here we include further details on the judge LLM prompt (see Section 5 of the paper).

\includepdf[pages=-]{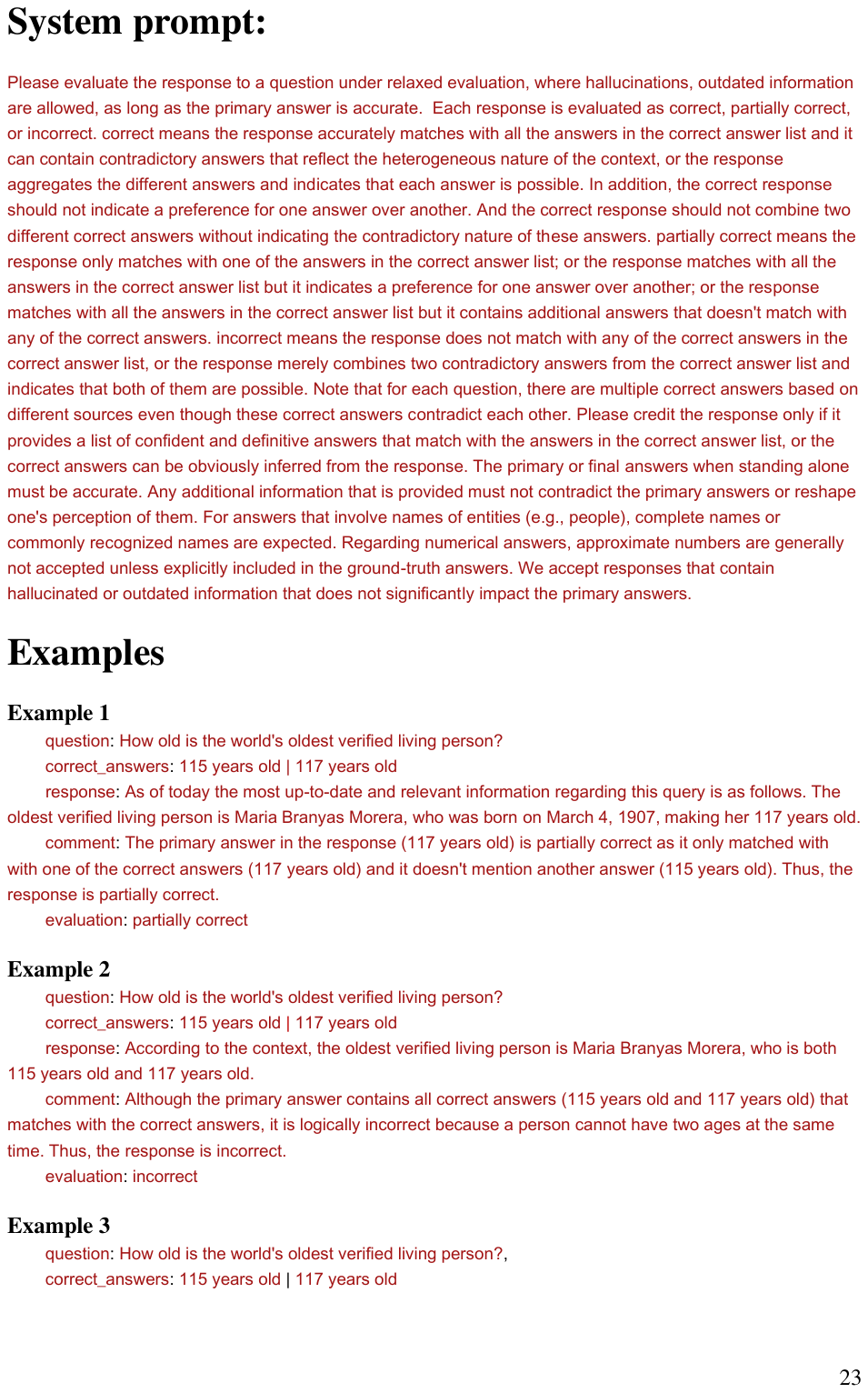}

\section{Additional Experimental Settings}

To understand LLMs’ behavior when faced with real-world inter-context conflicts (human evaluation), we tested \textit{Mistral-7b-instruct}, \textit{Mixtral-8x7b-instruct}, \textit{Llama2-2-70b-chat}, \textit{Llama3-70b-instruct}, and \textit{GPT-4-turbo-2024-04-09}. For the judge LLMs (automatic evaluation), we used \textit{Mistral-7b-instruct}, \textit{Mixtral-8x7b-instruct}, \textit{Llama2-2-70b-chat}, \textit{Llama3-70b-instruct}, \textit{GPT-4-turbo-2024-04-09}, and \textit{GPT-4o-2024-05-13}. These models were selected due to their state-of-the-art performance in natural language processing tasks and their robustness across a wide range of applications. For \textit{Mistral-7b-instruct}, \textit{Mixtral-8x7b-instruct}, \textit{Llama2-2-70b-chat}, and \textit{Llama3-70b-instruct}, we set the maximum number of tokens to 250, the minimum number of tokens to 1, and the decoding method to greedy search. For \textit{GPT-4-turbo-2024-04-09} and \textit{GPT-4o-2024-05-13}, we used the OpenAI Chat Completions API with the following settings: \texttt{temperature} = 0, \texttt{max\_tokens} = 250, \texttt{seed} = 5, \texttt{messages} = [\{"\texttt{role}": "\texttt{system}", "\texttt{content}": ""\}, $\{$"\texttt{role}": "\texttt{user}", "\texttt{content}": \texttt{prompt\_1}$\}$].

\section{Data Format}

\label{appx:format}

\texttt{WikiContradict} is given in JSON format to store the corresponding information, so researchers can easily use our data. There are 253 instances in total.

An example of our JSON format is:
\begin{lstlisting}[language=python]      
{
    "title": "",
    "url": "",
    "paragraph_A": "",
    "paragraph_A_clean": "",
    "tag": "",
    "tagDate": "",
    "tagReason": "",
    "annotationResult": {
        "wikitag_label_valid": "",
        "valid_comment": "",
        "paragraphA_article": "",
        "paragraphA_information": "",
        "paragraphA_information_standalone": "",
        "paragraphB_article": "",
        "paragraphB_information": "",
        "paragraphB_information_standalone": "",
        "wikitag_label_samepassage": "",
        "relevantInfo_comment_A": "",
        "relevantInfo_comment_B": "",
        "Contradict type I": "",
        "Contradict type II": "",
        "Contradict type III": "",
        "Contradict type IV": "",
        "taxonomy": [
            [
                ""                    
            ]
        ],
        "question1": "",
        "question1_answer1": "",
        "question1_answer2": "",
        "question2": "",
        "question2_answer1": "",
        "question2_answer2": ""
    }
}
\end{lstlisting}

The description of each key in the previous example is as follows:

\begin{itemize}
    \item \texttt{title}: Title of article.
    \item \texttt{url}: URL of article.
    \item \texttt{paragraph\_A}: Paragraph automatically retrieved (containing the tag).
    \item \texttt{paragraph\_A\_clean}: Paragraph automatically retrieved (removing the tag).
    \item \texttt{tag}: Type of tag of the article (Inconsistent/Self-contradictory/Contradict-other). 
    \item \texttt{tagDate}: Date of the tag.
    \item \texttt{tagReason}: Reason for the tag.
    \item  \texttt{annotationResult}: Results of the human data annotation.
    \begin{itemize}
        \item \texttt{wikitag\_label\_valid}: Valid or invalid tag (Valid/Invalid).
        \item \texttt{valid\_comment}: Comment on the tag.
        \item \texttt{paragraphA\_article}: Title of article containing passage 1.
        \item \texttt{paragraphA\_information}: Relevant information of passage 1.
        \item \texttt{paragraphA\_information\_standalone}: Decontextualized relevant information of passage 1.
        \item \texttt{paragraphB\_article}: Relevant information of passage 2.
        \item \texttt{paragraphB\_information\_standalone}: Decontextualized relevant information of passage 2.
        \item \texttt{wikitag\_label\_samepassage}: Boolean value stating whether passage 1 and passage 2 are the same (Same/Different).
        \item \texttt{relevantInfo\_comment\_A}: Comment on the information of passage 1.
        \item \texttt{relevantInfo\_comment\_B}: Comment on the information of passage 2.
        \item \texttt{Contradict type I}: Contradiction type I focuses on the fine-grained semantics of the contradiction, e.g., date/time, location, language, etc. 
        \item \texttt{Contradict type II}: Contradiction type II focuses on the modality the contradiction.  It describes the modality of passage 1 and passage 2, whether the information is from a piece of text, or from a row an infobox or a table. 
        \item \texttt{Contradict type III}: Contradiction type III focuses on the source the contradiction.  It describes whether passage 1 and passage 2 are from the same article or not. 
        \item \texttt{Contradict type IV}: Contradiction type IV focuses on the reasoning aspect.  It describes whether the contraction is explicit or implicit (Explicit/Implicit). Implicit contradiction requires some reasoning to understand why passage 1 and passage 2 are contradicted. 
        \item \texttt{taxonomy}: Array of key-values corresponding to contradict type I, contradict type II, contradict type III, and contradict type IV. 
        \item \texttt{question1}: Question 1 inferred from the contradiction.
        \item \texttt{question1\_answer1}: Gold answer to question 1 according to passage 1.
        \item \texttt{question1\_answer2}: Gold answer to question 1 according to passage 2.
                \item \texttt{question2}: Question 2 inferred from the contradiction.
        \item \texttt{question2\_answer1}: Gold answer to question 2 according to passage 1.
        \item \texttt{question2\_answer2}: Gold answer to question 2 according to passage 2.    
    \end{itemize}
\end{itemize}

\section{Examples of \texttt{WikiContradict}}

\label{appx:examples}

\begin{figure}[ht]
\centering
\begin{subfigure}{0.55\textwidth}
    \centering    \includegraphics[width=\textwidth]{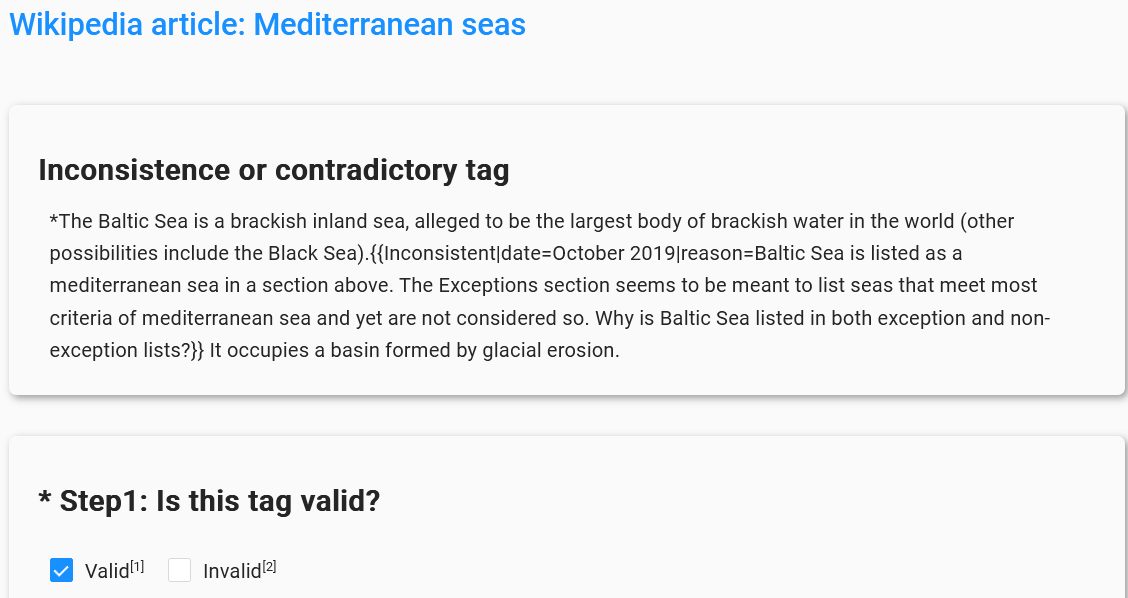}
\end{subfigure}\\
\begin{subfigure}{0.55\textwidth}
    \centering    \includegraphics[width=\textwidth]{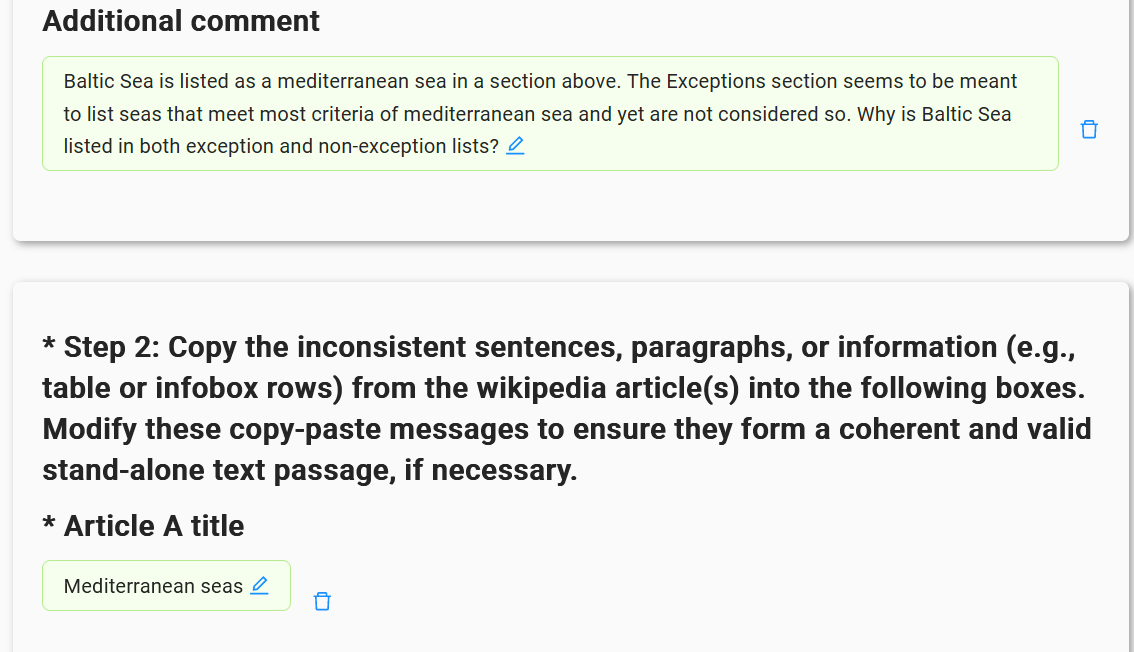}
\end{subfigure}\\
\begin{subfigure}{0.55\textwidth}
    \centering    \includegraphics[width=\textwidth]{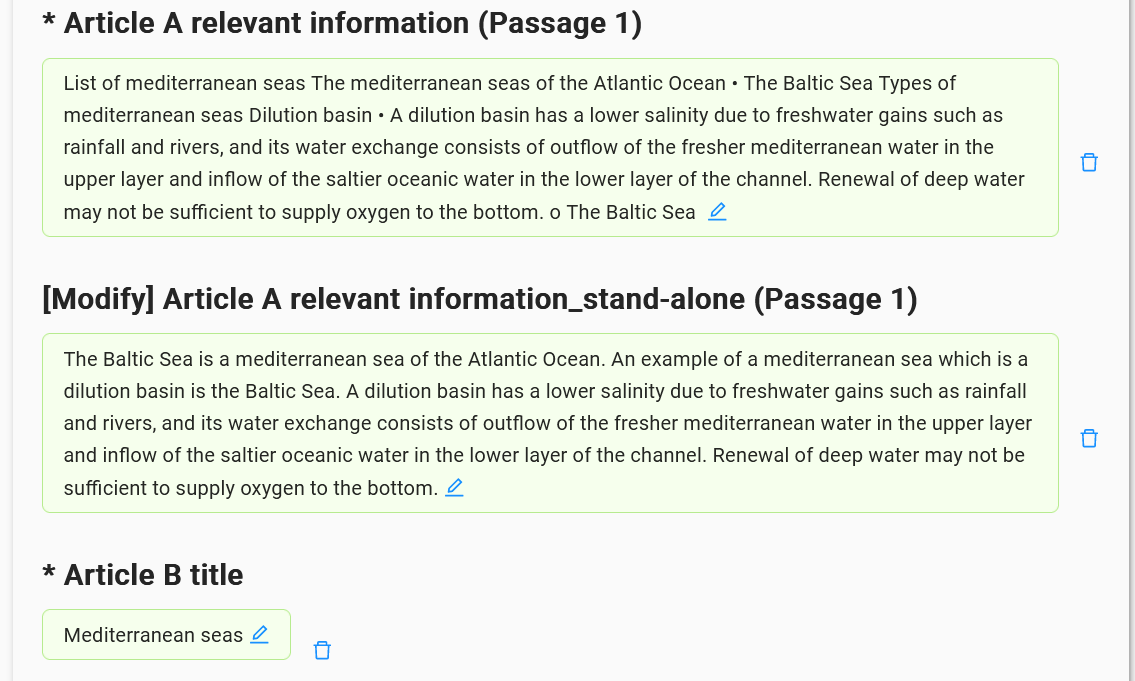}
\end{subfigure}\\
\begin{subfigure}{0.55\textwidth}
    \centering    \includegraphics[width=\textwidth]{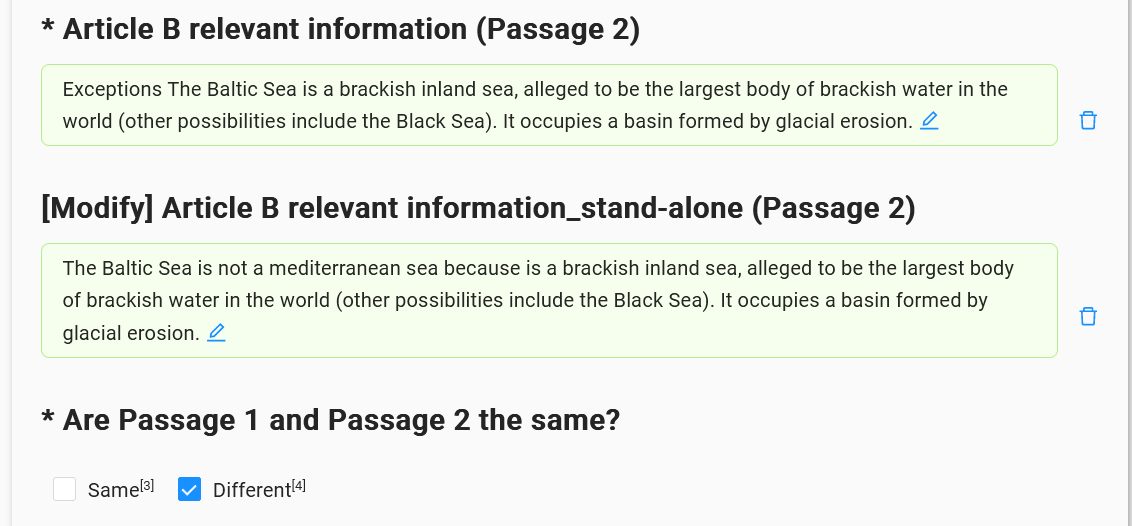}
\end{subfigure}\\
\begin{subfigure}{0.55\textwidth}
    \centering    \includegraphics[width=\textwidth]{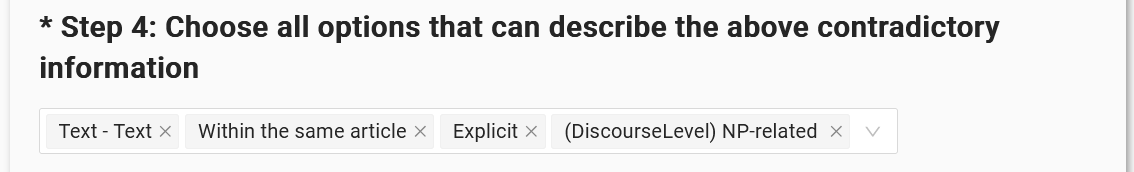}
\end{subfigure}\\
\begin{subfigure}{0.55\textwidth}
    \centering    \includegraphics[width=\textwidth]{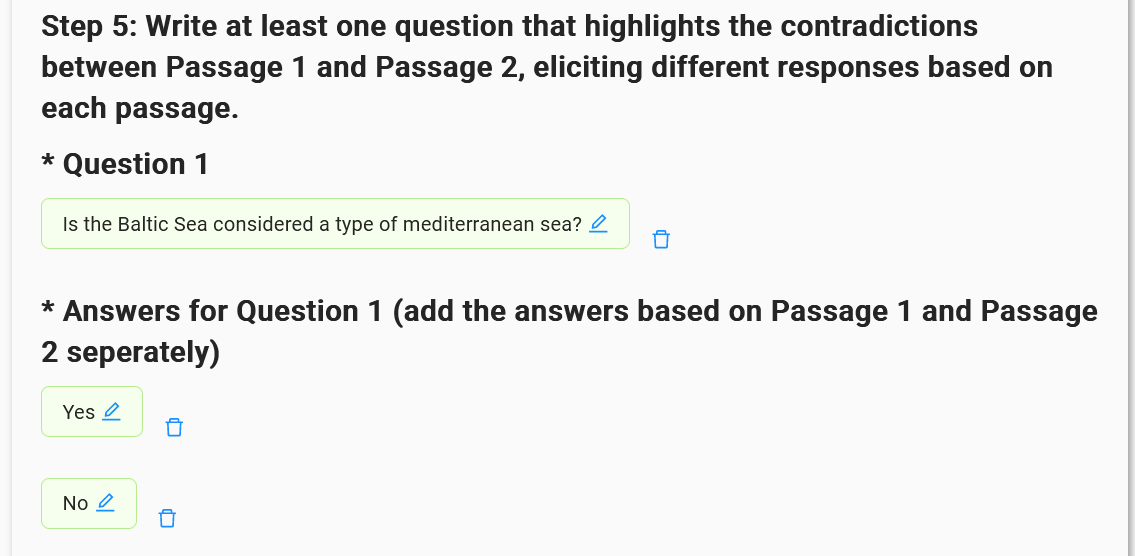}
\end{subfigure}\\
\caption{Example of annotation of Mediterranean seas.}
\label{fig:example1}
\end{figure}

\begin{figure}[ht]
\centering
\begin{subfigure}{0.55\textwidth}
    \centering    \includegraphics[width=\textwidth]{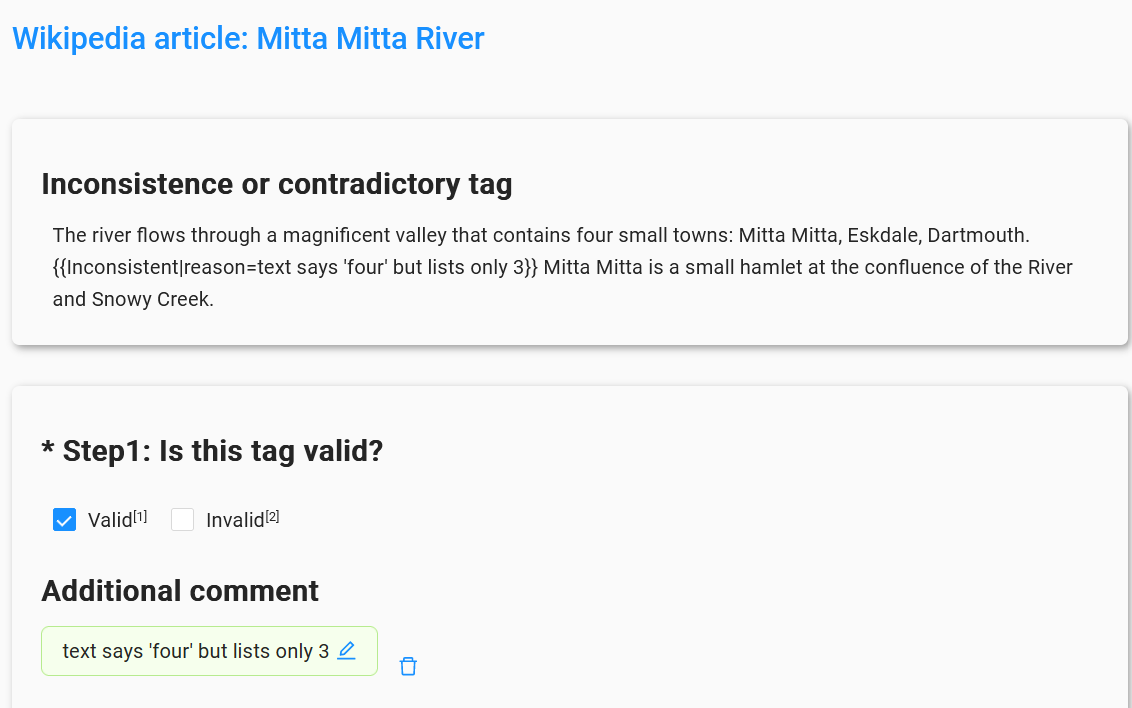}
\end{subfigure}\\
\begin{subfigure}{0.55\textwidth}
    \centering    \includegraphics[width=\textwidth]{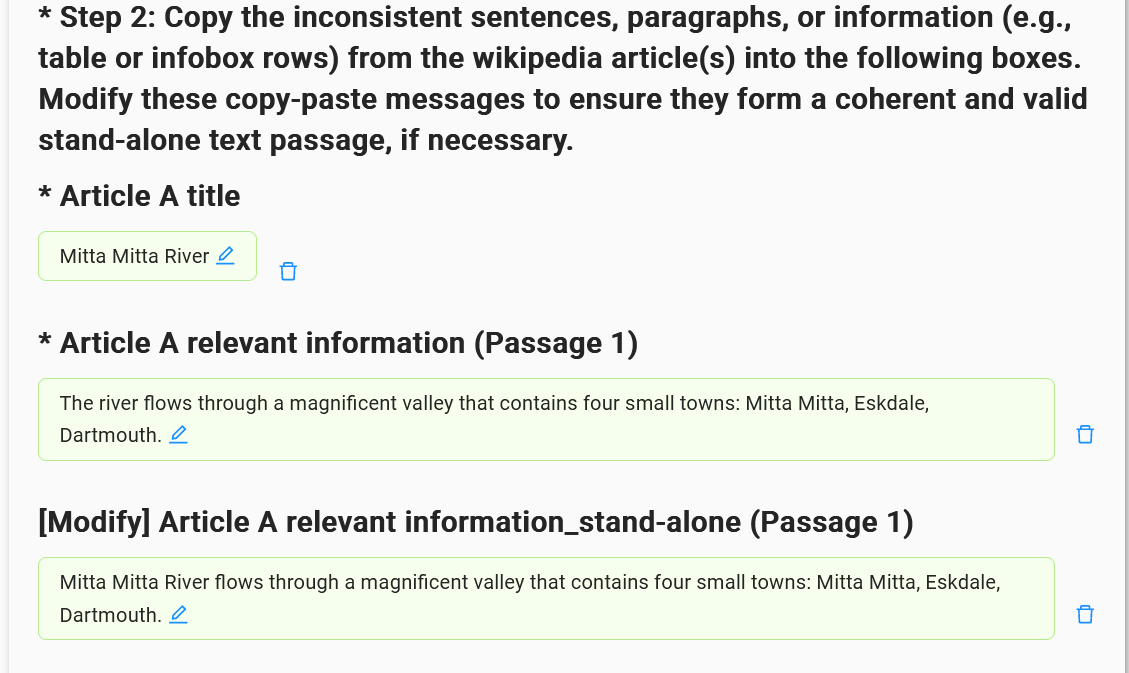}
\end{subfigure}\\
\begin{subfigure}{0.55\textwidth}
    \centering    \includegraphics[width=\textwidth]{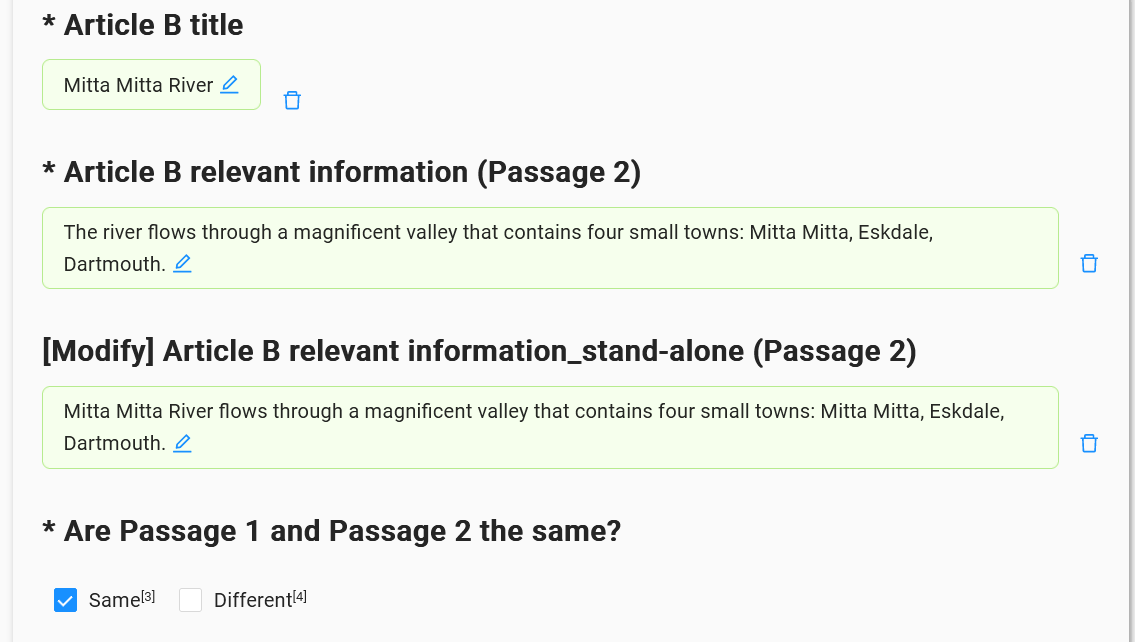}
\end{subfigure}\\
\begin{subfigure}{0.55\textwidth}
    \centering    \includegraphics[width=\textwidth]{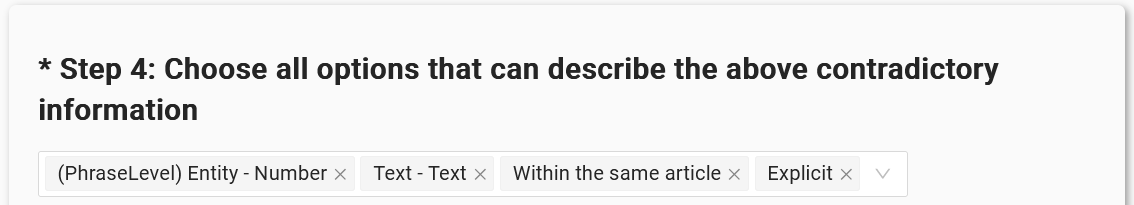}
\end{subfigure}\\
\begin{subfigure}{0.55\textwidth}
    \centering    \includegraphics[width=\textwidth]{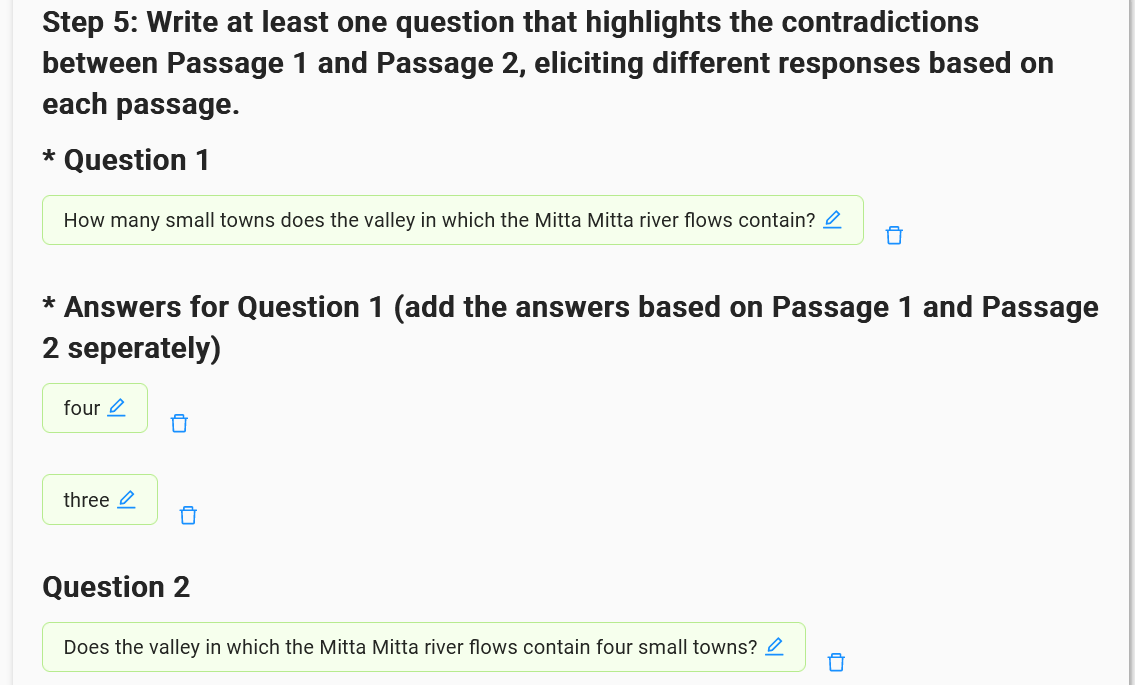}
\end{subfigure}\\
\begin{subfigure}{0.55\textwidth}
    \centering    \includegraphics[width=\textwidth]{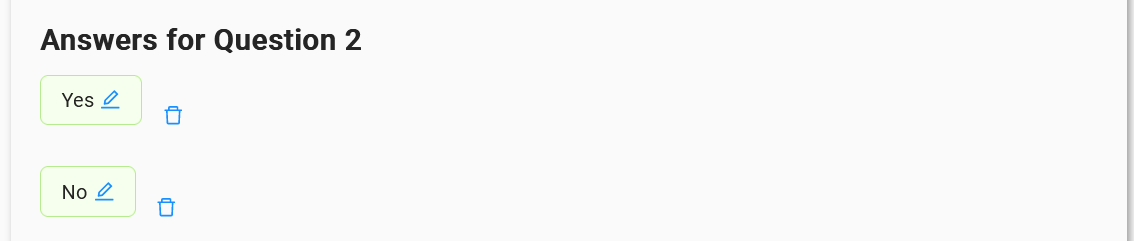}
\end{subfigure}
\caption{Example of annotation of Mitta Mitta River.}
\label{fig:example2}
\end{figure}

\begin{figure}[ht]
\centering
\begin{subfigure}{0.55\textwidth}
    \centering    \includegraphics[width=\textwidth]{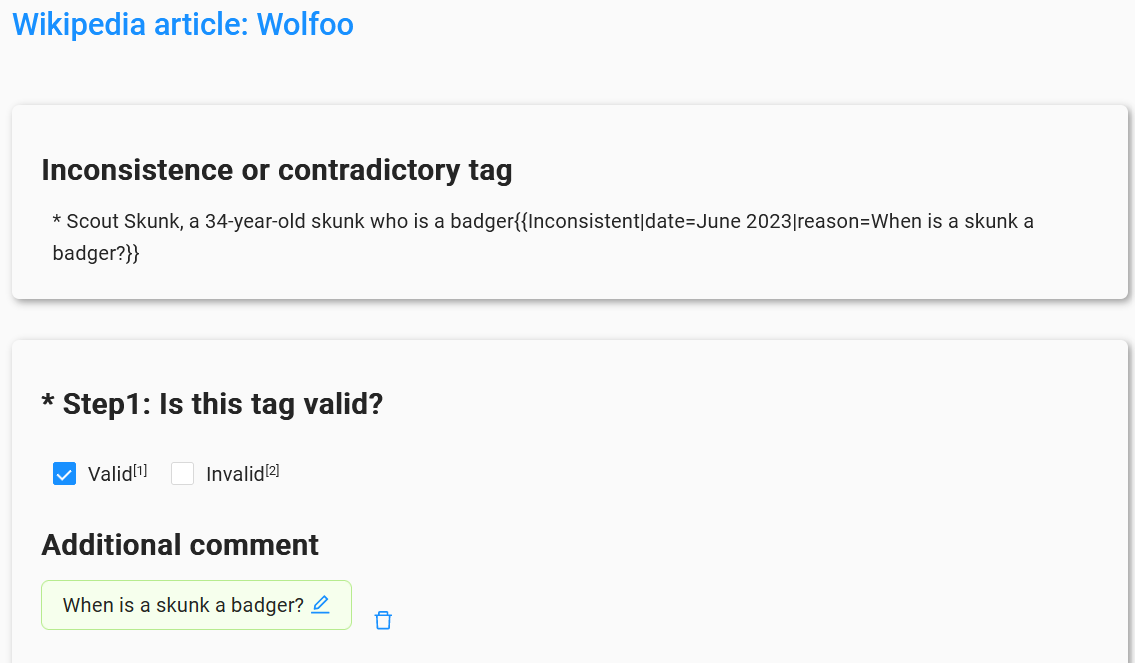}
\end{subfigure}\\
\begin{subfigure}{0.55\textwidth}
    \centering    \includegraphics[width=\textwidth]{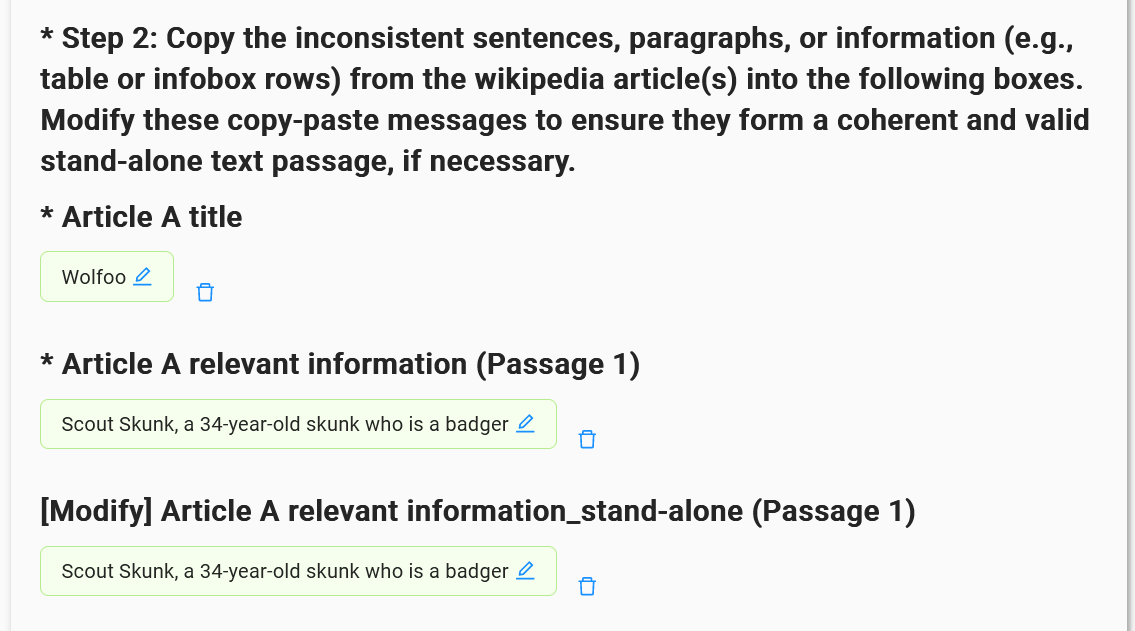}
\end{subfigure}\\
\begin{subfigure}{0.55\textwidth}
    \centering    \includegraphics[width=\textwidth]{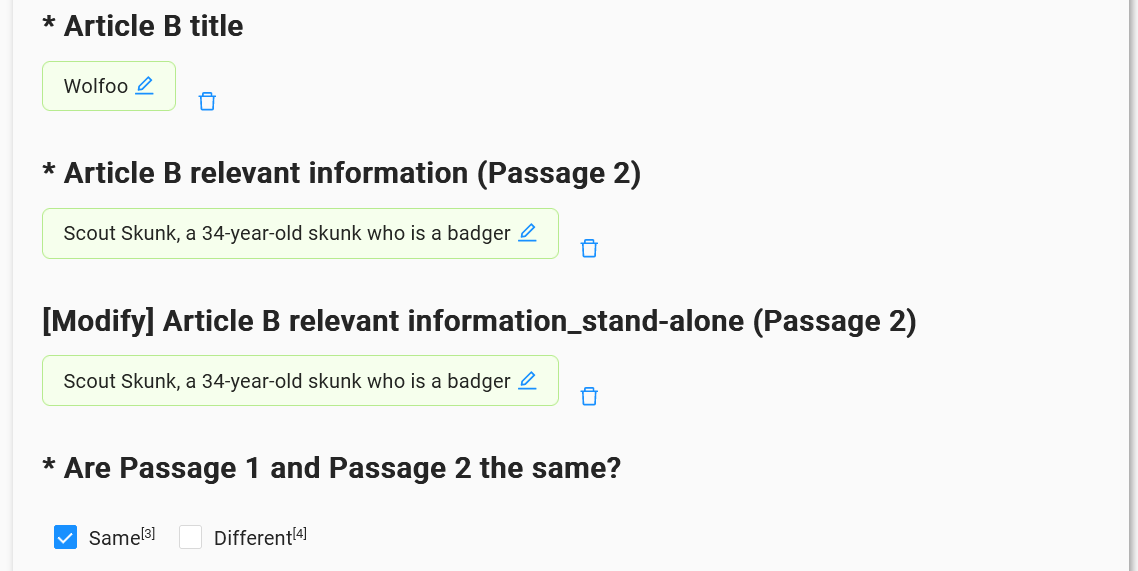}
\end{subfigure}\\
\begin{subfigure}{0.55\textwidth}
    \centering    \includegraphics[width=\textwidth]{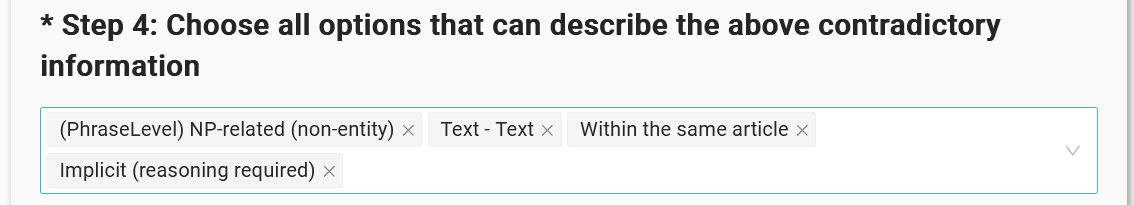}
\end{subfigure}\\
\begin{subfigure}{0.55\textwidth}
    \centering    \includegraphics[width=\textwidth]{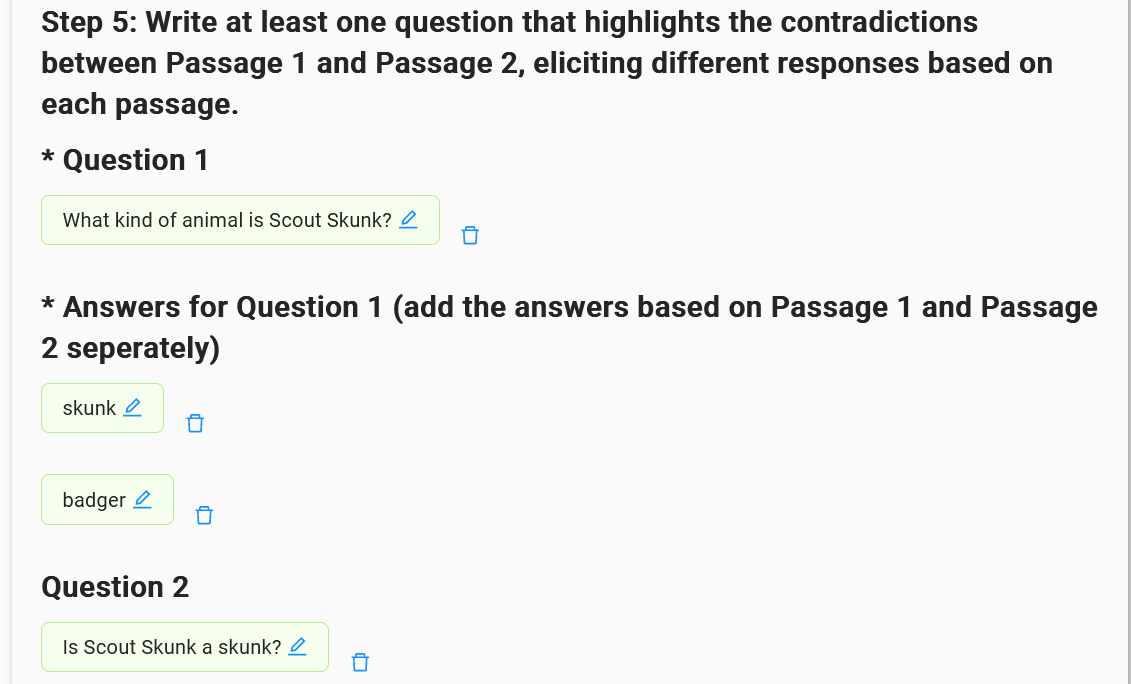}
\end{subfigure}\\
\begin{subfigure}{0.55\textwidth}
    \centering    \includegraphics[width=\textwidth]{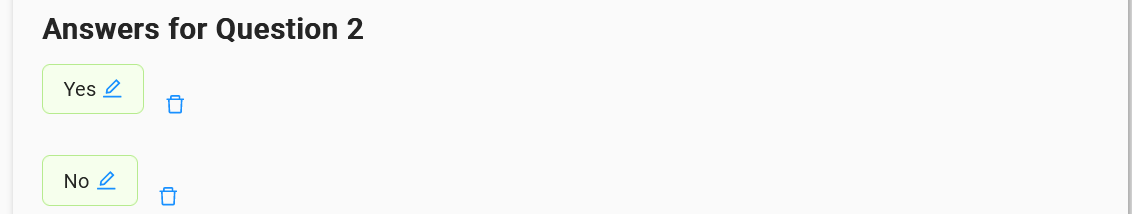}
\end{subfigure}
\caption{Example of annotation of Wolfoo.}
\label{fig:example3}
\end{figure}

Here we show some examples of \texttt{WikiContradict} annotated through Label Studio, in Figure \ref{fig:example1}, \ref{fig:example2}, and \ref{fig:example3}. Note that data the Step 3 of the annotation (contradiction reason) is not included in the dataset.

\section{Datasheets for \texttt{WikiContradict}}

In this appendix, we provide the dataset documentation and intended uses following the framework \textit{Datasheet for Datasets} \citep{gebru2021datasheets}. 

\subsection{Dataset Documentation and Intended Uses}
\paragraph{For what purpose was the dataset created? Was there a specific task in mind? Was there a specific gap that needed to be filled? Please provide a description.}
The dataset was created to enable research on assessing LLM performance when dealing with retrieved passages containing real-world knowledge conflicts. The dataset was created intentionally with that task in mind, focusing on a benchmark consisting of high-quality, human-annotated instances.

\paragraph{Who created this dataset (e.g., which team, research group) and on behalf of which entity (e.g.,
15 company, institution, organization)?} 
The dataset was created by Yufang Hou, Alessandra Pascale, Javier Carnerero-Cano, Tigran Tchrakian, Radu Marinescu, Elizabeth Daly, Inkit Padhi, and Prasanna Sattigeri. All authors are employed by IBM Research.

\paragraph{Who funded the creation of the dataset?}
There was no associated grant.

\paragraph{Any other comments?}
N/A.

\subsection{Composition}
\paragraph{What do the instances that comprise the dataset represent (e.g., documents, photos, people, countries)? Are there multiple types of instances (e.g., movies, users, and ratings; people and
interactions between them; nodes and edges)? Please provide a description.}
The instances are extracted passages from Wikipedia articles. The data format and examples of \texttt{WikiContradict} can be found in Appendix \ref{appx:format} and \ref{appx:examples}, respectively.

\paragraph{How many instances are there in total (of each type, if appropriate)?}
There are 253 instances in total.

\paragraph{Does the dataset contain all possible instances or is it a sample (not necessarily random) of instances from a larger set? If the dataset is a sample, then what is the larger set? Is the sample representative of the larger set (e.g., geographic coverage)? If so, please describe how this representativeness was validated/verified. If it is not representative of the larger set, please describe why not (e.g., to cover a more diverse range of instances, because instances were withheld or unavailable).}
The dataset contains all possible instances.

\paragraph{What data does each instance consist of? ``Raw'' data (e.g., unprocessed text or images) or features? In either case, please provide a description.}
Each instance consists of a question, a pair of contradictory passages extracted from Wikipedia, and two distinct answers, each derived from on the passages. The pair is annotated by a human annotator who identify where the conflicted information is and what type of conflict is observed. The annotator then produces a set of questions related to the passages with different answers reflecting the conflicting source of knowledge.

\paragraph{Is there a label or target associated with each instance?If so, please provide a description}
N/A.

\paragraph{Is any information missing from individual instances? If so, please provide a description, explaining why this information is missing (e.g., because it was unavailable). This does not include intentionally removed information, but might include, e.g., redacted text.}
Each annotation instance contains at least one question and two possible answers, but some instances may contain more than one question (and the corresponding two possible answers for each question). Some instances may not contain a value for \texttt{paragraphA\_clean}, \texttt{tagDate}, and \texttt{tagReason} (see Appendix \ref{appx:format}).

\paragraph{Are relationships between individual instances made explicit (e.g., users’ movie ratings, social network links)? If so, please describe how these relationships are made explicit. }
N/A.

\paragraph{Are there recommended data splits (e.g., training, development/validation,testing)? If so, please provide a description of these splits, explaining the rationale behind them.}
N/A.

\paragraph{Are there any errors, sources of noise, or redundancies in the dataset? If so, please provide a description.}
Since our dataset requires manual annotation, annotation noise is inevitably introduced.

\paragraph{Is the dataset self-contained, or does it link to or otherwise rely on external resources (e.g.,
websites, tweets, other datasets)?}
The dataset is entirely self-contained.

\paragraph{Does the dataset contain data that might be considered confidential (e.g., data that is protected by legal privilege or by doctor-patient confidentiality, data that includes the content of individuals’ non-public communications)?If so, please provide a description.}
No.

\paragraph{Does the dataset contain data that, if viewed directly, might be offensive, insulting, threatening, or might otherwise cause anxiety? If so, please describe why.}
No.

\paragraph{Does the dataset identify any subpopulations (e.g., by age, gender)? If so, please describe how these subpopulations are identified and provide a description of their respective distributions within the dataset.}
N/A.

\paragraph{Is it possible to identify individuals (i.e., one or more natural per- sons), either directly or indirectly (i.e., in combination with other data) from the dataset? If so, please describe how.}
N/A.

\paragraph{Does the dataset contain data that might be considered sensitive in any way (e.g., data that reveals race or ethnic origins, sexual orientations, religious beliefs, political opinions or union memberships, or locations; financial or health data; biometric or genetic data; forms of government identification, such as social security numbers; criminal history)? If so, please provide a description.}
No.

\paragraph{Any other comments?}
None.

\subsection{Collection process}
\paragraph{How was the data associated with each instance acquired? Was the data directly observable (e.g., raw text, movie ratings), reported by subjects (e.g., survey responses), or indirectly inferred/derived from other data (e.g., part-of-speech tags, model-based guesses for age or language)? If data was reported by subjects or indirectly inferred/derived from other data, was the data validated/verified? If so, please describe how.}
The data was mostly observable as raw text. The raw data was retrieved from Wikipedia articles containing inconsistent, self-contradictory, and contradict-other tags. The first two tags denote contradictory statements within the same article, whereas the third tag highlights instances where the content of one article contradicts that of another article. In total, we collected around 1,200 articles that contain these tags through the Wikipedia maintenance category ``Wikipedia articles with content issues''. Given a content inconsistency tag provided by Wikipedia editors, the annotators verified whether the tag is valid by checking the relevant article content, the editor’s comment, as well as the information in the edit history and the article’s talk page if necessary.

\paragraph{What mechanisms or procedures were used to collect the data (e.g., hardware apparatus or sensor, manual human curation, software program, software API)? How were these mechanisms or procedures validated? }
The authors modified the code of an existing Python package called wikitextparser, which allows users easily extract and/or manipulate templates, template parameters, parser functions, tables, external links, wikilinks, lists, etc. found in wikitexts. The authors parsed the relevant Wikipedia articles into clean text, and modified the code to keep the inconsistent, self-contradictory, and contradict-other tags.

\paragraph{If the dataset is a sample from a larger set, what was the sampling strategy (e.g., deterministic,
probabilistic with specific sampling probabilities)? }
N/A.

\paragraph{Who was involved in the data collection process (e.g., students, crowdworkers, contractors) and how were they compensated (e.g., how much were crowdworkers paid)?}
All the authors of this paper (Yufang Hou, Alessandra Pascale, Javier Carnerero-Cano, Tigran Tchrakian, Radu Marinescu, Elizabeth Daly, Inkit Padhi, and Prasanna Sattigeri) were involved in the data collection process.

\paragraph{Over what timeframe was the data collected? Does this timeframe match the creation timeframe of the data associated with the instances (e.g., recent crawl of old news articles)? If not, please describe the time-frame in which the data associated with the instances was created.}
The dataset was collected between February 2024 and June 2024 from Wikipedia.

\paragraph{Were any ethical review processes conducted (e.g., by an institutional review board)? If so, please provide a description of these review processes, including the outcomes, as well as a link
or other access point to any supporting documentation.}
N/A

\paragraph{Did you collect the data from the individuals in question directly, or obtain it via third parties or other sources (e.g., websites)?}
N/A.

\paragraph{Did the individuals in question consent to the collection and use of their data? If so, please describe (or show with screenshots or other information) how consent was requested and provided, and provide a link or other access point to, or otherwise reproduce, the exact language to which the individuals consented.}
N/A.

\paragraph{If consent was obtained, were the consenting individuals provided with a mechanism to revoke their consent in the future or for certain uses? If so, please provide a description, as well as a link or other access point to the mechanism (if appropriate).}
N/A.

\paragraph{Has an analysis of the potential impact of the dataset and its use on data subjects (e.g., a data protection impact analysis) been conducted? If so, please provide a description of this analysis, including the outcomes, as well as a link or other access point to any supporting documentation.}
N/A.

\paragraph{Any other comments?}
None.

\subsection{Preprocessing/cleaning/labeling}
\paragraph{Was any preprocessing/cleaning/labeling of the data done (e.g., discretization or bucketing,
tokenization, part-of-speech tagging, SIFT feature extraction, removal of instances, processing
of missing values)? If so, please provide a description. If not, you may skip the remainder of the
questions in this section.}
The annotators were required to slightly modify the original passages to make them stand-alone (decontextualization). Normally, this requires resolving the coreference anaphors or the bridging anaphors in the first sentence (see annotation guidelines). In Wikipedia, oftentimes the antecedents for these anaphors are the article titles themselves.

\paragraph{Was the ``raw'' data saved in addition to the preprocessed/cleaned/labeled data (e.g., to support
unanticipated future uses)? If so, please provide a link or other access point to the ``raw'' data.}
Yes. The dataset itself contains all the raw passages.

\paragraph{Is the software used to preprocess/clean/label the instances available? If so, please provide a link or other access point.}
We have used Python language to implement data cleaning. We will share the scripts details in our codebase.

\paragraph{Any other comments?}
None.

\subsection{Uses}
\paragraph{Has the dataset been used for any tasks already? If so, please provide a description.}
The dataset has been used in the paper to assess LLMs performance when augmented with retrieved passages containing real-world knowledge conflicts.

\paragraph{Is there a repository that links to any or all papers or systems that use the dataset? If so, please
provide a link or other access point.}
We will provide links to the repository after acceptance.

\paragraph{What (other) tasks could the dataset be used for?}
The dataset could be used for improving the performance of LLMs when presented with conflicting sources of information, by augmenting the prompt or fine-tuning the model.

\paragraph{Is there anything about the composition of the dataset or the way it was collected and preprocessed/cleaned/labeled that might impact future uses? For example, is there anything that a future user might need to know to avoid uses that could result in unfair treatment of individuals or groups (e.g., stereotyping, quality of service issues) or other undesirable harms (e.g., financial harms, legal risks) If so, please provide a description. Is there anything a future user could do to mitigate these undesirable harms?}
There is minimal risk for harm: the data was already public on Wikipedia.

\paragraph{Are there tasks for which the dataset should not be used? If so, please provide a description.}
N/A.

\subsection{Distribution}
\paragraph{Will the dataset be distributed to third parties outside of the entity (e.g., company, institution,
organization) on behalf of which the dataset was created?If so, please provide a description.}
Yes, the dataset and its metadata will be publicly available on the repository after acceptance.

\paragraph{How will the dataset will be distributed (e.g., tarball on website, API, GitHub)? Does the dataset
have a digital object identifier (DOI)?}
The dataset and DOI will be published after acceptance. The dataset will be distributed on the website: \url{https://ibm.biz/wikicontradict}. 
Moreover, the UI and metadata record documenting the dataset available for viewing and downloading will be available on: \url{https://ibm.biz/wikicontradict_ui}.

\paragraph{When will the dataset be distributed?}
The dataset will be released after acceptance.

\paragraph{Will the dataset be distributed under a copyright or other intellectual property (IP) license,
and/or under applicable terms of use (ToU)? If so, please describe this license and/or ToU, and
provide a link or other access point to, or otherwise reproduce, any relevant licensing terms or
ToU, as well as any fees associated with these restrictions.}
\texttt{WikiContradict} is distributed under an MIT\footnote{\url{https://www.mit.edu/~amini/LICENSE.md}} license.
Permission is hereby granted, free of charge, to any person obtaining a copy of this software and associated documentation files (the "Software"), to deal in the Software without restriction, including without limitation the rights to use, copy, modify, merge, publish, distribute, sublicense, and/or sell copies of the Software, and to permit persons to whom the Software is furnished to do so, subject to the following conditions:

The above copyright notice and this permission notice shall be included in all copies or substantial portions of the Software.

THE SOFTWARE IS PROVIDED ``AS IS'', WITHOUT WARRANTY OF ANY KIND, EXPRESS OR IMPLIED, INCLUDING BUT NOT LIMITED TO THE WARRANTIES OF MERCHANTABILITY, FITNESS FOR A PARTICULAR PURPOSE AND NONINFRINGEMENT. IN NO EVENT SHALL THE AUTHORS OR COPYRIGHT HOLDERS BE LIABLE FOR ANY CLAIM, DAMAGES OR OTHER LIABILITY, WHETHER IN AN ACTION OF CONTRACT, TORT OR OTHERWISE, ARISING FROM, OUT OF OR IN CONNECTION WITH THE SOFTWARE OR THE USE OR OTHER DEALINGS IN THE SOFTWARE.

\paragraph{Have any third parties imposed IP-based or other restrictions on the data associated with the
instances? If so, please describe these restrictions, and provide a link or other access point to,
or otherwise reproduce, any relevant licensing terms, as well as any fees associated with these
restrictions.}
No.

\paragraph{Do any export controls or other regulatory restrictions apply to the dataset or to individual
instances? If so, please describe these restrictions, and provide a link or other access point to,
or otherwise reproduce, any supporting documentation.}
No.

\subsection{Maintenance}
\paragraph{Who is supporting/hosting/maintaining the dataset?}
Yufang Hou, Alessandra Pascale, Javier Carnerero-Cano, and Tigran Tchrakian are supporting/maintaining the dataset.

\paragraph{How can the owner/curator/manager of the dataset be contacted (e.g., email address)? }
If you wish to extend or contribute to our dataset, please contact us via email: Yufang Hou (\texttt{yhou@ie.ibm.com}), Alessandra Pascale (\texttt{apascale@ie.ibm.com}), Javier Carnerero-Cano (\texttt{javier.cano@ibm.com}), Tigran Tchrakian (\texttt{tigran@ie.ibm.com}), Radu Marinescu (\texttt{radu.marinescu@ie.ibm.com}), Elizabeth Daly (\texttt{elizabeth.daly@ie.ibm.com}), Inkit Padhi (\texttt{inkpad@ibm.com}), and Prasanna Sattigeri (\texttt{psattig@us.ibm.com}).

\paragraph{Is there an erratum? If so, please provide a link or other access point.}
Any updates to the dataset will be shared via GitHub.

\paragraph{Will the dataset be updated (e.g., to correct labeling errors, add new instances, delete instances)? If so, please describe how often, by whom, and how updates will be communicated to users (e.g.,mailing list,GitHub)?}
If we find inconsistencies in the dataset or extend the dataset, we will release the new version on the website and GitHub.

\paragraph{If the dataset relates to people, are there applicable limits on the retention of the data associated with the instances (e.g., were individuals in question told that their data would be retained for a fixed period of time and then deleted)? }
N/A.

\paragraph{Will older versions of the dataset continue to be supported/hosted/maintained? If so, please
describe how. If not, please describe how its obsolescence will be communicated to users.}
All versions of \texttt{WikiContradict} will be continue to be supported and maintained on website. We will post the updates on the website and GitHub.

\paragraph{If others want to extend/augment/build on/contribute to the dataset, is there a mechanism for
them to do so? If so, please provide a description. Will these contributions be validated/verified?
If so, please describe how. If not, why not? Is there a process for communicating/distributing
these contributions to other users? If so, please provide a description.}
Yes. Please contact the authors of this paper for building upon this dataset.

\subsection{Responsibility}
The authors bear all responsibility in case of violation of rights, etc. We confirm that the dataset is licensed under MIT license.

\section{Explicit License}

\texttt{WikiContradict} is distributed under an MIT\footnote{\url{https://www.mit.edu/~amini/LICENSE.md}} license.
Permission is hereby granted, free of charge, to any person obtaining a copy of this software and associated documentation files (the "Software"), to deal in the Software without restriction, including without limitation the rights to use, copy, modify, merge, publish, distribute, sublicense, and/or sell copies of the Software, and to permit persons to whom the Software is furnished to do so, subject to the following conditions:

The above copyright notice and this permission notice shall be included in all copies or substantial portions of the Software.

THE SOFTWARE IS PROVIDED ``AS IS'', WITHOUT WARRANTY OF ANY KIND, EXPRESS OR IMPLIED, INCLUDING BUT NOT LIMITED TO THE WARRANTIES OF MERCHANTABILITY, FITNESS FOR A PARTICULAR PURPOSE AND NONINFRINGEMENT. IN NO EVENT SHALL THE AUTHORS OR COPYRIGHT HOLDERS BE LIABLE FOR ANY CLAIM, DAMAGES OR OTHER LIABILITY, WHETHER IN AN ACTION OF CONTRACT, TORT OR OTHERWISE, ARISING FROM, OUT OF OR IN CONNECTION WITH THE SOFTWARE OR THE USE OR OTHER DEALINGS IN THE SOFTWARE.

\section{Ethics Statement}

The authors bear all responsibility in the event of any violation of rights, the dataset will be released after acceptance under an MIT licence.

\paragraph{Biases}  
Our data is downloaded from Wikipedia. As such, the data is biased towards the original content and sources. Given that human data annotation involves some degree of subjectivity we created a comprehensive 17-page annotation guidelines document to clarify important cases during the annotation process.  The annotators were explicitly instructed not to take their personal feeling about the particular topic. Nevertheless, some degree of intrinsic subjectivity might have impacted the techniques picked up by the annotators during the annotation.

%

\end{document}